\documentclass{article}

\PassOptionsToPackage{numbers, compress}{natbib}

\usepackage[final]{neurips_2024}

\usepackage[acronym,nowarn,section,nogroupskip,nonumberlist]{glossaries}

\glsdisablehyper{}
\newcommand{\acaps}[1]{{\scshape #1}}
\newacronym{bo}{\acaps{bo}}{Bayesian Optimization}
\newacronym{mobo}{\acaps{mobo}}{Multi-Objective Bayesian Optimization}
\newacronym{gp}{\acaps{gp}}{Gaussian process}
\newacronym{plm}{\acaps{plm}}{Pretrained Language Model}
\newacronym{nlp}{\acaps{nlp}}{Natural Language Processing}
\newacronym{nlu}{\acaps{nlu}}{Natural Language Understanding}
\newacronym{nlg}{\acaps{nlg}}{Natural Language Generation}
\newacronym{mlm}{\acaps{mlm}}{Masked Language Model}
\newacronym{swa}{\acaps{swa}}{Stochastic Weight Averaging}
\newacronym{sam}{\acaps{sam}}{Sharpness-Aware Minimization}
\newacronym{de}{\acaps{de}}{Deep Ensemble}
\newacronym{mcc}{\acaps{mcc}}{Matthews Correlation Coefficient}
\newacronym{pcc}{\acaps{pcc}}{Pearson Correlation Coefficient}
\newacronym{lora}{\acaps{l}o\acaps{ra}}{Low-Rank Adaptation}
\newacronym{qlora}{\acaps{ql}o\acaps{ra}}{Quantized Low-Rank Adaptation}
\newacronym{t5}{\acaps{t5}}{Text-to-Text Transfer Transformer}

\newacronym{llm}{\acaps{llm}}{Large Language Model}

\newacronym{ehvi}{\acaps{ehvi}}{Expected HyperVolume Improvement}

\newacronym{ours}{\acaps{bomf}}{Bayesian Optimization Model Fusion}

\newacronym{robertabase}{RoBERTa-base}{RoBERTa-base}
\glsunset{robertabase}

\newacronym{llamasevenb}{\texttt{Llama-7B}}{Llama-7B}
\glsunset{llamasevenb}

\newacronym{llama27b}{\acaps{ll}a\acaps{ma2}-\acaps{7b}}{Llama2-7B}
\glsunset{llama27b}

\newacronym{llama38b}{\acaps{ll}a\acaps{ma3}-\acaps{8b}}{Llama3-8B}
\glsunset{llama38b}

\newacronym{bert}{\acaps{bert}}{Bidirectional Encoder Representations from Transformers}
\glsunset{bert}

\newacronym{gpt}{\acaps{gpt}}{Generative Pre-trained Transformer}
\glsunset{gpt}

\newacronym{roberta}{\acaps{r}o\acaps{bert}a}{Robustly Optimized BERT Pretraining Approach}
\glsunset{roberta}

\newacronym{llama}{\acaps{ll}a\acaps{ma}}{Large Language Model Meta AI}
\glsunset{llama}

\newacronym{mrpc}{\acaps{mrpc}}{}
\usepackage{amsmath, amssymb, amsthm}
\usepackage{mathtools}
\usepackage{bbm}
\usepackage{dsfont}

\newcommand{\br}{\mathbf{r}}

\newcommand{\bw}{\mathbf{w}} 
\newcommand{\bx}{\mathbf{x}}

\newcommand{\bW}{\mathbf{W}}

\newcommand{\bY}{\mathbf{Y}}

\newcommand{\calD}{{\mathcal{D}}}

\newcommand{\calH}{{\mathcal{H}}}

\newcommand{\calL}{{\mathcal{L}}}
\newcommand{\calM}{{\mathcal{M}}}

\newcommand{\calP}{{\mathcal{P}}}

\newcommand{\calS}{{\mathcal{S}}}

\newcommand{\bbR}{\mathbb{R}}

\newcommand{\bdelta}{{\boldsymbol{\delta}}}

\newcommand{\btheta}{{\boldsymbol{\theta}}}

\newcommand{\blambda}{{\boldsymbol{\lambda}}}

\theoremstyle{plain}

\theoremstyle{definition}

\theoremstyle{remark}

\DeclareMathOperator*{\argmax}{arg\,max}
\DeclareMathOperator*{\argmin}{arg\,min}

\def\[#1\]{\begin{equation}\begin{aligned}#1\end{aligned}\end{equation}}

\newcommand{\floss}{f_{\textrm{loss}}}
\newcommand{\fmetric}{f_{\textrm{metric}}}

\usepackage[usenames,dvipsnames]{xcolor}

\usepackage{graphicx}
\usepackage{subcaption}
\usepackage{algorithm}
\usepackage{algorithmic}

\usepackage{booktabs, array}

\definecolor{citecolor}{RGB}{0,102,204}
\definecolor{linkcolor}{RGB}{190,105,30}
\definecolor{urlcolor}{RGB}{199,21,133}

\usepackage[colorlinks,linktoc=all,pagebackref=true]{hyperref}
\usepackage[all]{hypcap}
\hypersetup{citecolor=citecolor}
\hypersetup{linkcolor=linkcolor}
\hypersetup{urlcolor=urlcolor}
\usepackage[nameinlink,capitalise, noabbrev]{cleveref}
\creflabelformat{equation}{#2\textup{#1}#3}  
\crefname{section}{\S}{\S\S}

\newsavebox\CBox 

\def\UL#1{\underline{#1}}
\def\BL#1{\sbox\CBox{#1}\resizebox{\wd\CBox}{\ht\CBox}{\underline{\textbf{#1}}}}

\usepackage[utf8]{inputenc} 
\usepackage[T1]{fontenc}    
\usepackage{hyperref}       
\usepackage{url}            
\usepackage{booktabs}       
\usepackage{amsfonts}       
\usepackage{nicefrac}       
\usepackage{microtype}      
\usepackage{xcolor}         

\usepackage{comment}
\usepackage{wrapfig}
\usepackage{colortbl}
\usepackage{multirow}
\usepackage{mathtools}
\usepackage{enumitem}
\usepackage{makecell}
\usepackage{kotex}
\usepackage{tabularx}

\title{Model Fusion through Bayesian Optimization in Language Model Fine-Tuning}

\author{%
    Chaeyun Jang\thanks{Co-first authors} \\
    KAIST \\
    \texttt{jcy9911@kaist.ac.kr} \\
    \And
    Hyungi Lee\footnotemark[1] \\
    KAIST \\
    \texttt{lhk2708@kaist.ac.kr} \\
    \AND
    Jungtaek Kim\thanks{Co-corresponding authors} \\
    University of Pittsburgh \\
    \texttt{jungtaek.kim@pitt.edu} \\
    \And
    Juho Lee\footnotemark[2] \\
    KAIST \\
    \texttt{juholee@kaist.ac.kr} \\
}

\begin{document}

\maketitle

\begin{abstract}
Fine-tuning pre-trained models for downstream tasks is a widely adopted technique known for its adaptability and reliability across various domains. Despite its conceptual simplicity, fine-tuning entails several troublesome engineering choices, such as selecting hyperparameters and determining checkpoints from an optimization trajectory. To tackle the difficulty of choosing the best model, one effective solution is \emph{model fusion}, which combines multiple models in a parameter space. However, we observe a \emph{large discrepancy between loss and metric landscapes} during the fine-tuning of pre-trained language models. Building on this observation, we introduce a novel model fusion technique that optimizes both the desired metric and loss through \emph{multi-objective Bayesian optimization}. In addition, to effectively select hyperparameters, we establish a two-stage procedure by integrating Bayesian optimization processes into our framework. Experiments across various downstream tasks show considerable performance improvements using our Bayesian optimization-guided method. Code will be available at: \url{https://github.com/chaeyoon-jang/bomf.git}.
\end{abstract}

\section{Introduction}
\label{main:sec:introduction}

The field of \gls{nlp} has significantly advanced with the pre-training of Transformer-based models on large amounts of texts without supervision.
In general, these pre-trained networks are fine-tuned on supervised downstream tasks to solve particular tasks. The rise of \glspl{llm} such as \gls{gpt}~\citep{radford2018improving} and \gls{llama}~\citep{touvron2023llama} has increased demands for huge memory and computing during fine-tuning on downstream tasks.
In response, low rank approximation methods such as \gls{lora}~\citep{hu2021lora} and \gls{qlora}~\citep{qlora2023dettmers}
have been adopted recently to fine-tune the \gls{llm}.
However, the fine-tuning of \glspl{plm} exhibits high sensitivity to marginal variations in hyperparameters such as learning rate and batch size,
often leading to training failure and the performance drop of a fine-tuned model~\citep{mosbach2020stability},
while searching hyperparameters requires a vast amount of resources.

An effective strategy to seek an optimal model among multiple candidates is model ensembling, which yields impressive performance on generalization and robustness~\citep{lakshminarayanan2017simple}. 
However, traditional ensemble methods lead to several drawbacks including the space and computational costs that linearly scale with the number of models involved.
These issues are particularly pertinent for \glspl{llm},
since individual models are costly to train and test.
Therefore,
we can alternatively utilize model fusion to aggregate multiple models into a single proficient model on a parameter space.
One of its simplest forms, known as \gls{swa}~\citep{izmailovaveraging}, involves taking the average of model parameters obtained during an optimization process. Despite its simplicity, \gls{swa} and its variants have proven successful across various tasks, notably in computer vision~\citep{izmailovaveraging,maddox2019simple,cha2021swad,nam2023decoupled}. Recent advancement in this field is the concept of Model Soups, which has been introduced by~\citet{wortsman2022model}.
This approach weight-averages a set of models, obtained from multiple fine-tuning runs with different hyperparameters to create a powerful model that outperforms both individual and ensemble models.

The effectiveness of model fusion has predominantly been explored in the visual domain.
For instance, while Model Soups have shown considerable improvements in image classification, they have not demonstrated superiority in the \gls{nlp} tasks~\citep{wortsman2022model}.
The existing averaging methods like \gls{swa} make use of their ability to encourage a fused model to locate on the center of the flatter area near local optima~\citep{izmailovaveraging,he2019asymmetric},
as loss landscapes are analogous to metric landscapes in computer vision tasks.
Unfortunately, for \glspl{plm}, loss landscapes are substantially mismatched to metric landscapes, so that the flat loss minimum found by \gls{swa} does not necessarily correspond to the flat metric minimum making a simple averaging method fail to find a superior model.

In this paper, we present a novel model fusion approach with an efficient hyperparameter selection strategy, denoted as \gls{ours}, specifically designed to fine-tune \glspl{plm}. To motivate our method, we start by illustrating two empirical analyses. Firstly, we demonstrate that the existing model fusion techniques are not suitable for \glspl{plm}. Secondly, we highlight that the optimal hyperparameters for \glspl{plm} exhibit consistency on varying the number of frozen layers or the rank used in the \gls{lora} setting~\citep{hu2021lora}.

Based on these findings, we introduce our proposed method to build a single model, aggregated through the weighted combination of individual models. Supposing that evaluation metrics are non-differentiable, we employ \gls{bo}~\citep{BrochuE2010arxiv, GarnettR2023book}, which is a black-box optimization technique,
in developing our model fusion method.
To the best of our knowledge, this is the first study that utilizes \gls{bo} in the context of model fusion,
in order to achieve the following objectives:
\begin{itemize}[leftmargin=20pt]
    \item \textbf{Utilization of Both Metrics and Loss Functions in Model Fusion.}
Instead of running \gls{bo} with an averaged target metric, we use \gls{mobo} that considers both metrics and loss functions for model fusion. Despite low correlations between loss and metric values, we find that incorporating loss values still serves as useful guidance.
    \item \textbf{Two-Stage Model Fusion.}
We devise our model fusion process as a two-stage \gls{bo} procedure. One is for optimizing hyperparameters in fine-tuning and the other is dedicated to our model fusion method. The objective of the first stage is to maximize gains from the second stage to find hyperparameters leading to the optimal fused model after the \gls{bo} of the second stage.
\end{itemize}

We demonstrate the effectiveness of \gls{ours} on several \gls{nlp} tasks, including both \gls{nlu} and \gls{nlg}, with \gls{roberta}, \gls{t5} and \gls{llama}. Through these comprehensive experiments, we assess the performance of \gls{ours} in diverse \gls{nlp} tasks and uncover the interesting properties of our approach through various ablation studies.
\section{Preliminaries}
\label{main:sec:back}

\paragraph{Problem Setup.}
In this paper, we explore the process of fine-tuning \glspl{plm} using two types of datasets: a downstream training dataset $\calD_{\text{trn}}$ and a validation dataset $\calD_{\text{val}}$. Assuming that we are given a \emph{pre-trained} set of weights $\btheta_{\text{init}}$ and a \emph{trainable} set of weights $\bw_{\text{init}}$ for our \gls{plm} denoted as $\calM(\btheta, \bw)$, $\bw_{\text{init}}$ is either a subset of $\btheta_{\text{init}}$ or \gls{lora} weights~\citep{hu2021lora}. Specifically, in the former case, $\bw_{\text{init}}$ is deliberately selected from $\btheta_{\text{init}}$. As a special case, $\bw_{\text{init}}$ will be identical to $\btheta_{\text{init}}$ if any layers or weights are not frozen. Meanwhile, if the \gls{lora} method is employed during the fine-tuning of our model, $\bw_{\text{init}}$ will be the \gls{lora} weights.

We use $K$ distinct metrics, denoted as $\fmetric^{(k)}(\calM, \calD)$ for $k \in [K]$, to evaluate our model's performance on a given task. Each metric $\fmetric^{(k)}$ is typically non-differentiable, while a differentiable loss function $\floss$ is employed for training. Assuming that $\calD_{\text{val}}$ is similar to the true data distribution, our goal is to find the optimal set of trainable weights $\bw^\star$ that minimizes the following:
\begin{align}
\label{eq:true}
\bw^\star=\argmin_{\bw\in\bW}\sum_{k=1}^K \Bar{f}_{\text{metric}}^{(k)}(\calM(\btheta_{\text{init}},\bw), \calD_{\text{val}}),
\end{align}
where $\Bar{f}_{\text{metric}}^{(k)}$ is a normalized version of $\fmetric^{(k)}$ for all $k\in[K]$, and $\bW$ is the space of trainable weights. However, due to the non-differentiability of the $\fmetric^{(k)}$ functions, conventional approaches resort to finding approximate solutions using gradient descent or its variants, as shown below:
\begin{align}
\Tilde{\bw}^\star=\argmin_{\bw \in \bW}\floss(\calM(\btheta_{\text{init}},\bw), \calD_{\text{trn}}).
\end{align}
As will be discussed in the subsequent section, misalignment between loss and metric surfaces is more prominent in \glspl{plm} compared to computer vision models. To address this challenge, we propose a novel method to more adequately make use of $\{\fmetric^{(k)}\}_{k=1}^K$ and $\calD_{\text{val}}$ by considering~\cref{eq:true}.

\paragraph{Model Fusion.}
In the recent work~\citep{izmailovaveraging,wortsman2022model,rame2023model,rame2023rewardedsoups}, there has been a growing interest in the use of \emph{weight averaging} or \emph{model fusion} across diverse tasks. This line of research is an effective strategy to achieve superior performance in downstream tasks, all while managing computational costs by aggregation of multiple models. In this context, the aggregation of multiple models involves the identification of a set of $N$ fine-tuned trainable weights, denoted as $\calS=\{\bw_i\}_{i=1}^N$. The objective is to derive a fused weight vector, $\Bar{\bw}$, by utilizing $\calS$, such that $\Bar{\bw}$ outperforms all other members in $\calS$. This can be expressed as $L_{\text{metric}}(\Bar{\bw})\leq \argmin_{\bw\in \calS}L_{\text{metric}}(\bw)$, where $L_{\text{metric}}(\bw)\coloneqq\sum_{k=1}^K \Bar{f}_{\text{metric}}^{(k)}(\calM(\btheta_{\text{init}},\bw),\calD_{\text{val}})$.

Model fusion approaches can be categorized into two main types: 1) uniform averaging and 2) weighted averaging. Uniform averaging methods, e.g., \gls{swa}~\citep{izmailovaveraging}, Greedy Soups~\citep{wortsman2022model}, involve the straightforward process of uniformly averaging weights within a subset $\Bar{\calS}\subseteq \calS$ to obtain an improved performing weight vector $\Bar{\bw}$, i.e. $\Bar{\bw} = \frac{1}{|\Bar{\calS}|}\sum_{\bw\in\Bar{\calS}}\bw$. Here, selecting a suitable subset $
\Bar{\calS}$ is an important strategy for each method. On the other hand, weighted averaging approaches, e.g., Learned Soups~\citep{wortsman2022model} and Rewarded Soups~\citep{rame2023rewardedsoups}, aim to determine an optimized weight vector $\bw$ by forming a convex combination of parameters from $\calS$, expressed as $\Bar{\bw} = \sum_{i=1}^N \delta_i\bw_i$, where each averaging coefficient $\delta_i$ satisfies $\delta_i \geq 0$, and $\sum_{i=1}^N \delta_i=1$. While weighted averaging methods offer more flexibility compared to uniform averaging, they often require additional training to determine suitable values for the coefficient set $\bdelta$ through gradient descent updates based on the loss function $\floss$. However, in our proposed method, we suggest a weighted averaging technique that considers not only the loss function $\floss$ but also the metrics $\{\fmetric^{(k)}\}_{k=1}^K$.

\paragraph{Multi-Objective Bayesian Optimization.}
\gls{bo} is a sample-efficient black-box optimization technique with probabilistic regression.
Since we assume that an objective to optimize is unknown,
a surrogate function, which is generally a probabilistic regression model, is estimated instead.
The key desired properties of the surrogate function are attained by considering how a search space is exploited and explored through its outputs.
Utilizing the surrogate function,
\gls{bo} eventually optimizes a specific form of optimizable function, called an acquisition function;
see~\citep{BrochuE2010arxiv,GarnettR2023book} for details.

On top of generic~\gls{bo},
\gls{mobo} is used to solve an optimization problem,
involved with $K$ different competing objectives:
\begin{equation}
    \bx^\dagger = \argmin_{\bx} (f_1(\bx), f_2(\bx), \ldots, f_K(\bx)).
    \label{eq:mobo}
\end{equation}
Supposing that we cannot directly access $f_1, f_2, \ldots, f_K$,
probabilistic surrogate models, which are alternatives to unknown objectives,
should be used to determine a next point to evaluate.
To find a solution candidate of~\cref{eq:mobo} using~\gls{mobo},
we can consider scalarization of either the realizations of surrogate models or acquisition functions corresponding to multiple objectives~\citep{PariaB2019uai}.
In contrast to the scalarization method, the maximization of \gls{ehvi}, on a metric space~\citep{EmmerichMTM2006ieeetec} can be used:
\begin{equation}
    \bx^{\dagger} = \argmax_{\bx}\, \textrm{EHVI}(\bx; \bY, \br),
    \label{ehvi}
\end{equation}
where a hypervolume is defined as the size of space between the Pareto frontier of $n$ historical evaluations $\bY \in \bbR^{n \times K}$ and a reference point $\br$.
While the scalarization determines query points by aggregating $K$ outputs with particular (potentially random) coefficients,
the hypervolume improvement maximization chooses query points that widen the expected hypervolume,
which is more robust to function scales without the sampling distributions of scalarization coefficients.
As reported in the previous work~\citep{DaultonS2021neurips,ament2023unexpected},
compared to other existing \gls{mobo} algorithms,
$q$NEHVI which is a variant of the EHVI method that evaluates a batch of $q$ points in a parallel manner. Building on the powerful \gls{mobo} algorithm, our model fusion framework is capable of determining averaging coefficients efficiently reducing the number of evaluations required to find better fused \glspl{plm}.
\section{Empirical Findings}
\label{main:sec:mismatch}

In this section, we present empirical observations motivating our model fusion strategy. In~\cref{main:subsec:landscape}, we initially illustrate distinct findings: unlike in computer vision tasks, in \gls{nlp} tasks, there exists a significant misalignment between the loss and metric surfaces. This misalignment poses a challenge for straightforward model fusion methods when fine-tuning \glspl{plm}. In~\cref{main:subsec:hyperparam_alignment}, we find that the optimal fine-tuned hyperparameters for \glspl{plm} analogously align across different architectural configurations varying the number of frozen layers or variations in rank in the \gls{lora} setting.

\subsection{On Misalignment in Loss and Metric Landscapes}
\label{main:subsec:landscape}

\begin{figure}[t]
    \centering
    \begin{subfigure}[b]{0.48\textwidth}
        \centering
        \includegraphics[width=\textwidth]{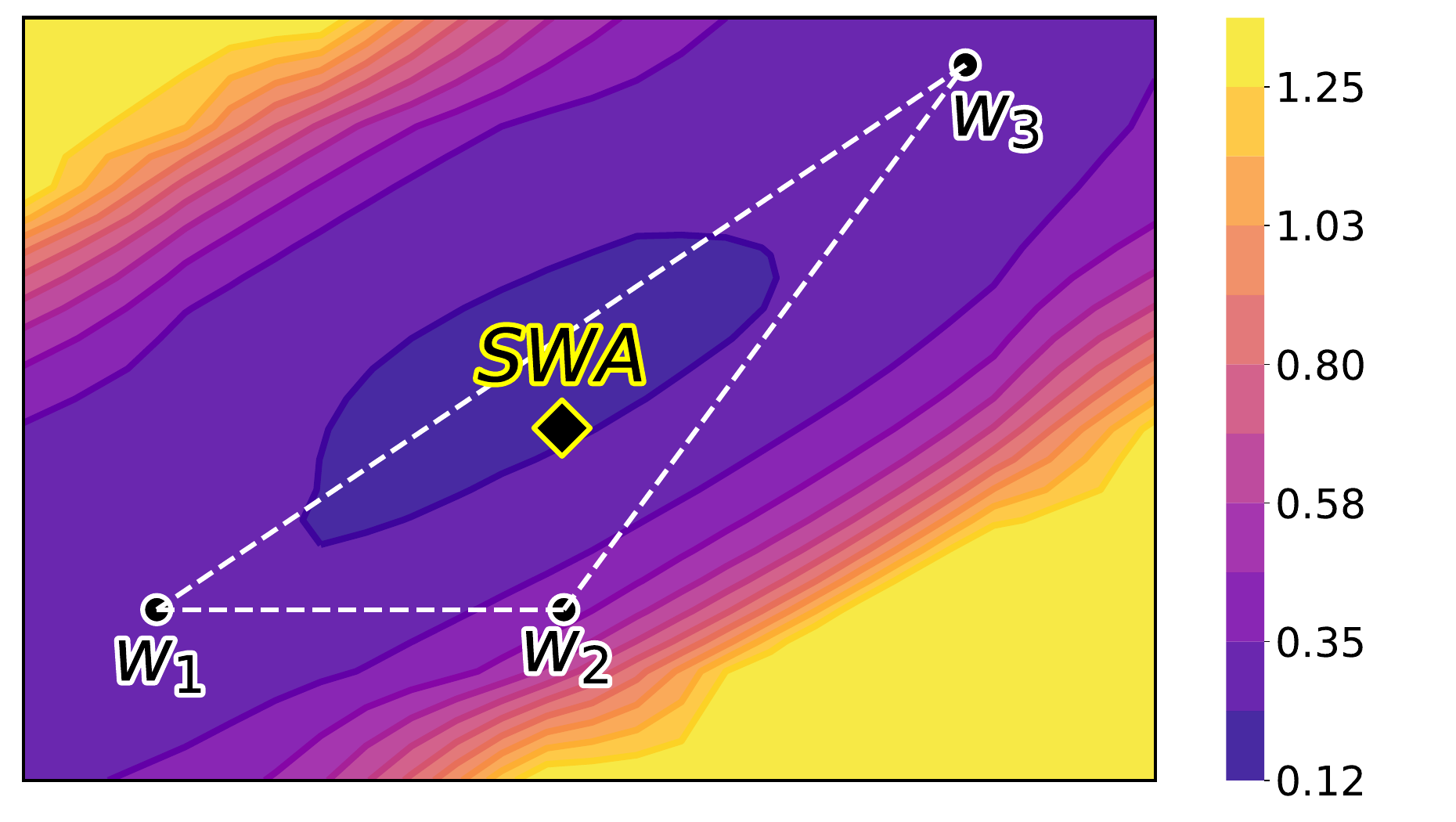}
        \caption{Loss of ResNet-50}
        \label{main:fig:vision_loss}
    \end{subfigure}
    \hfill
    \begin{subfigure}[b]{0.48\textwidth}
        \centering
        \includegraphics[width=\textwidth]{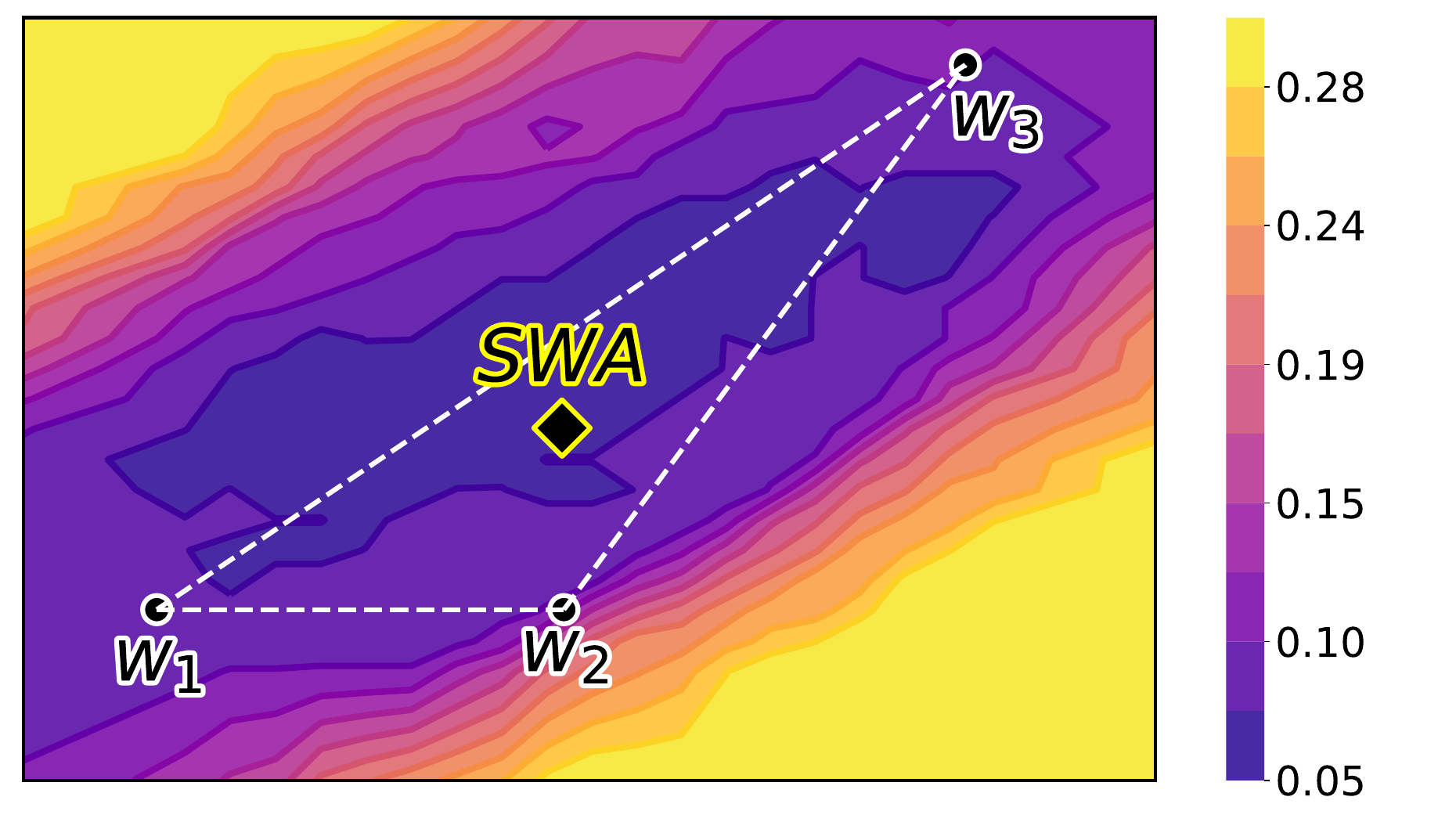}
        \caption{Metric of ResNet-50}
        \label{main:fig:vision_metric}
    \end{subfigure}
    \begin{subfigure}[b]{0.48\textwidth}
        \centering
        \includegraphics[width=\textwidth]{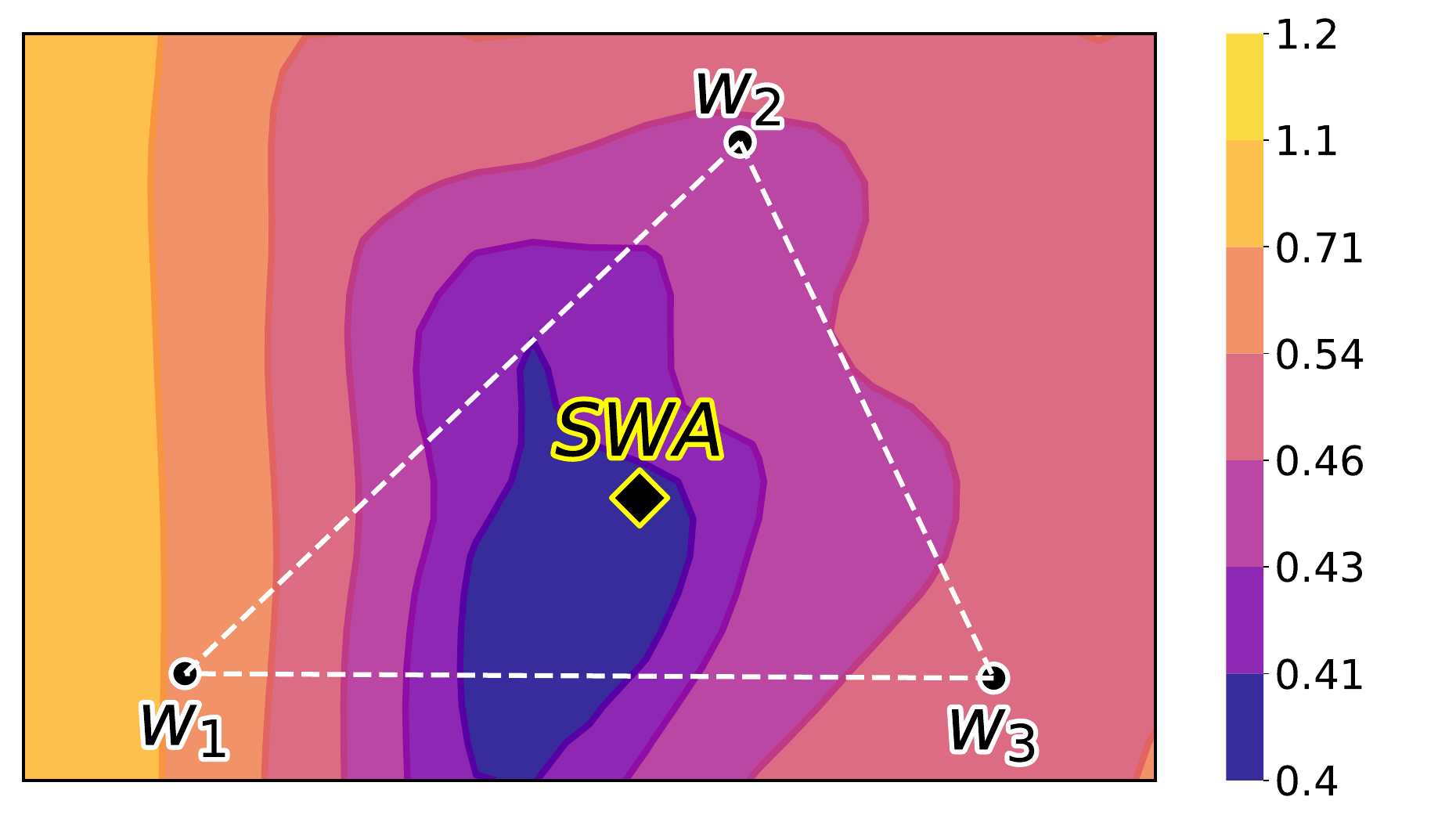}
        \caption{Loss of \gls{roberta}}
        \label{main:fig:language_loss}
    \end{subfigure}
    \hfill
    \begin{subfigure}[b]{0.48\textwidth}
        \centering
        \includegraphics[width=\textwidth]{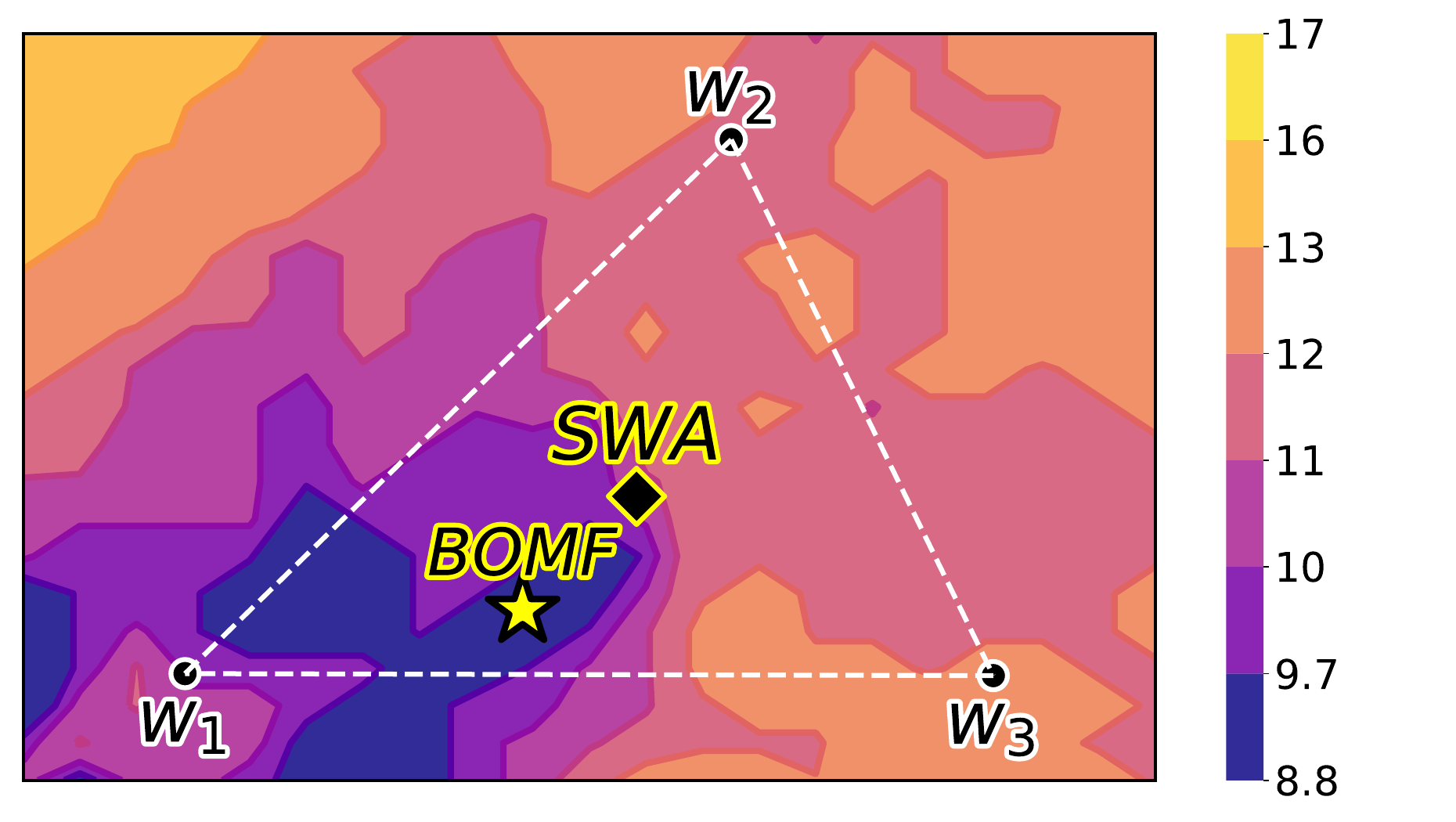}
        \caption{Metric of \gls{roberta}}
        \label{main:fig:language_metric}
    \end{subfigure}
    \caption{Visualization of the loss landscape over parameters (\cref{main:fig:vision_loss,main:fig:language_loss}) and the metric landscape over parameters (\cref{main:fig:vision_metric,main:fig:language_metric}) for both the vision task (\cref{main:fig:vision_loss,main:fig:vision_metric}) and the \gls{nlp} task (\cref{main:fig:language_loss,main:fig:language_metric}). The metric is $1-$accuracy and F1 score for the vision task and the \gls{nlp} task, respectively. In the vision task, we fine-tune the ResNet-50 model~\citep{he2016deep} pre-trained with ImageNet-21k~\citep{russakovsky2015imagenet} on the Caltech-101 dataset~\citep{li_andreeto_ranzato_perona_2022}, while in the \gls{nlp} task, fine-tuning was performed on the pre-trained \gls{roberta} model on the \gls{mrpc} dataset. The members of the \gls{swa} for each figure are denoted as $w_1, w_2, w_3$.}
    \label{fig:loss_metric_surfaces}
\end{figure}

The well-known success of uniform averaging, e.g., \gls{swa} and Model Soups, in image classification tasks, has been grounded on the flatness of a loss landscape. As one can see in~\cref{main:fig:vision_loss},
the use of uniform averaging successfully explores minima on the flatter region of the loss landscape using individual weights close to the flatter region, resulting in enhanced generalization loss on a test dataset. This generalization effect is similarly observed in the case of the metric landscape, as illustrated in \cref{main:fig:vision_metric}, owing to the similarity between the loss and metric landscapes. This similarity is the consequence of the inherent similarity between the loss function and the metric~\citep{mao2023cross}. However, the domain of language modeling, characterized by semantic, morphosyntactic, and pragmatic intricacies, requires the evaluation of generalization performance across a diverse array of tasks and metrics~\citep{dodge-etal-2019-show}.
It is unlikely to precisely align these metrics with a training loss function~\citep{zhukov2017differentiable,liu2022don},
leading to a misalignment that often results in more complex and less flat surfaces in language tasks compared to the loss function visually demonstrated in~\cref{main:fig:language_loss,main:fig:language_metric}.

In~\cref{main:fig:language_loss,main:fig:language_metric}, we find that while uniform averaging can reach high generalization performance based on the loss function, it poorly performs concerning the metric function compared to the best-performing weight in $\calS$. However, \cref{main:fig:language_metric} shows that even though the uniform averaging of three weight points degrades the metric performance, better points in terms of higher metric values exist in the convex set of the three weight points. The empirical results we observe above, which are caused by the complex and misaligned surface, motivate the need to utilize weighted averaging methods and seek the optimal combination of averaging weights based on the metric. This does not agree with the previous findings in vision tasks~\citep{wortsman2022model} and \cref{main:fig:vision_metric} which argue minimal performance difference between the weighted averaging and the uniform averaging. Refer to \cref{app:sec:additional_experiment_surface} for numerical results that show the discrepancy between the loss and metric landscapes in \glspl{plm}.

\subsection{On Hyperparameter Alignment}
\label{main:subsec:hyperparam_alignment}

\begin{figure}[t]
    \centering
    \includegraphics[width=0.48\textwidth]{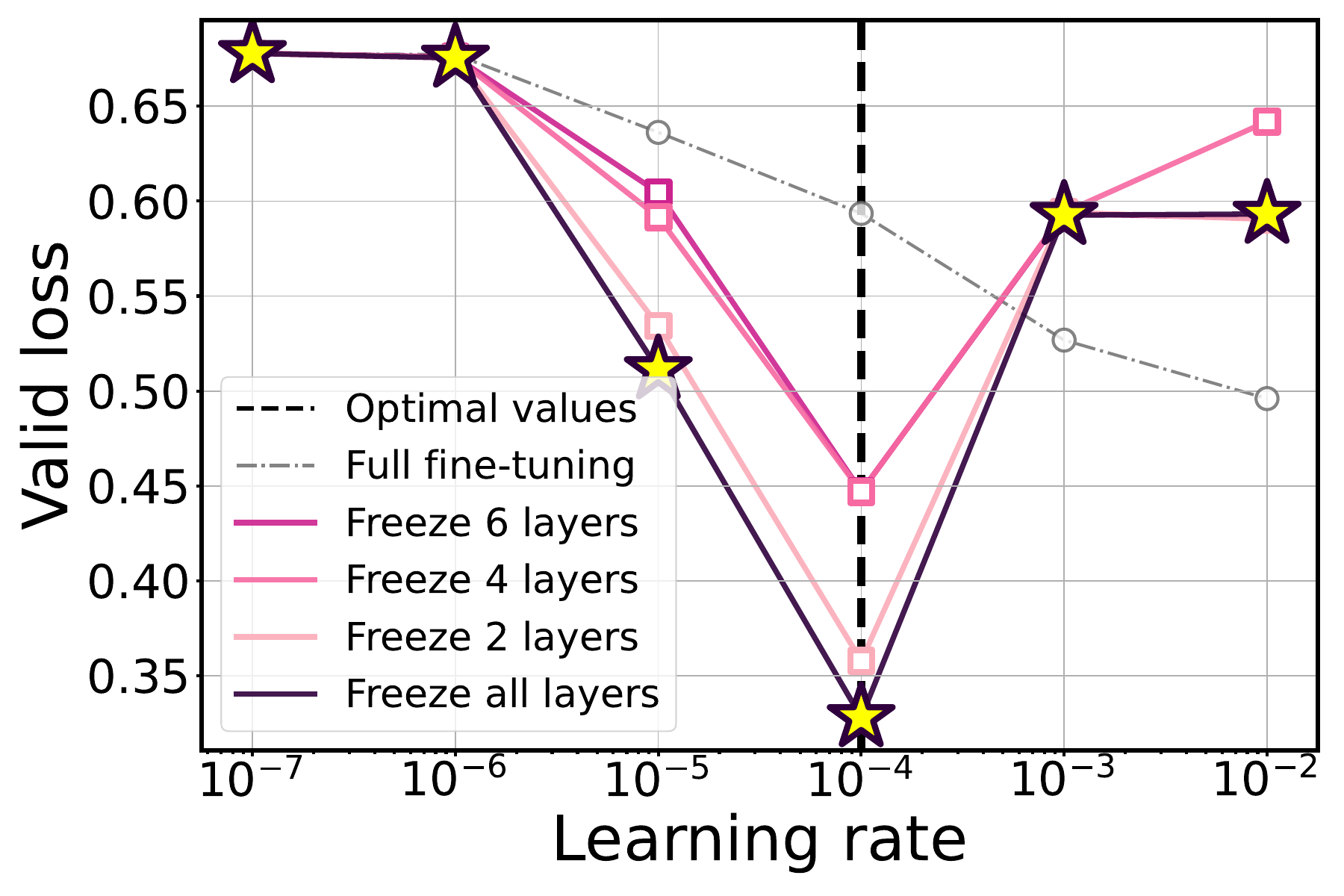}
    \includegraphics[width=0.48\textwidth]{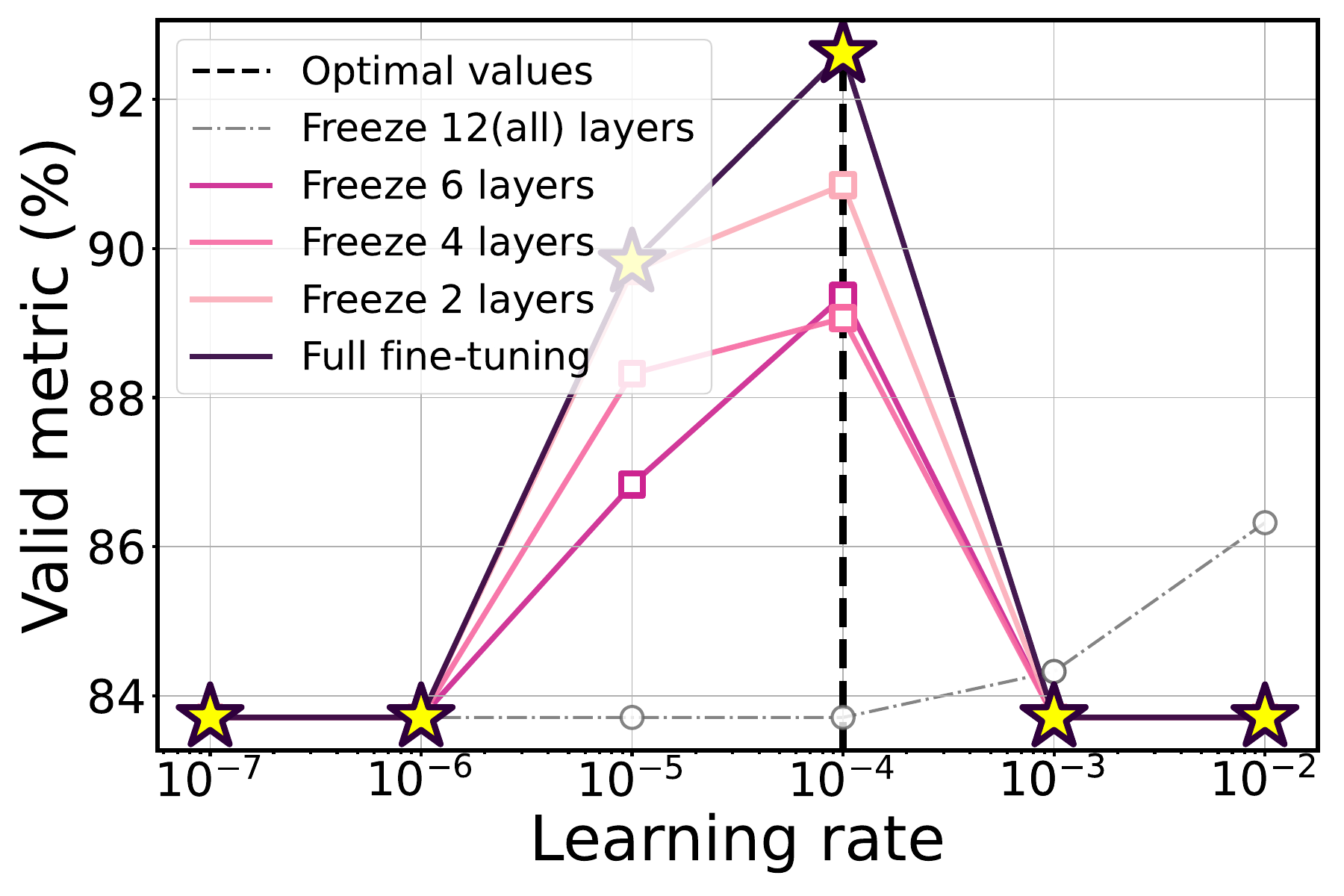}
    \includegraphics[width=0.48\textwidth]{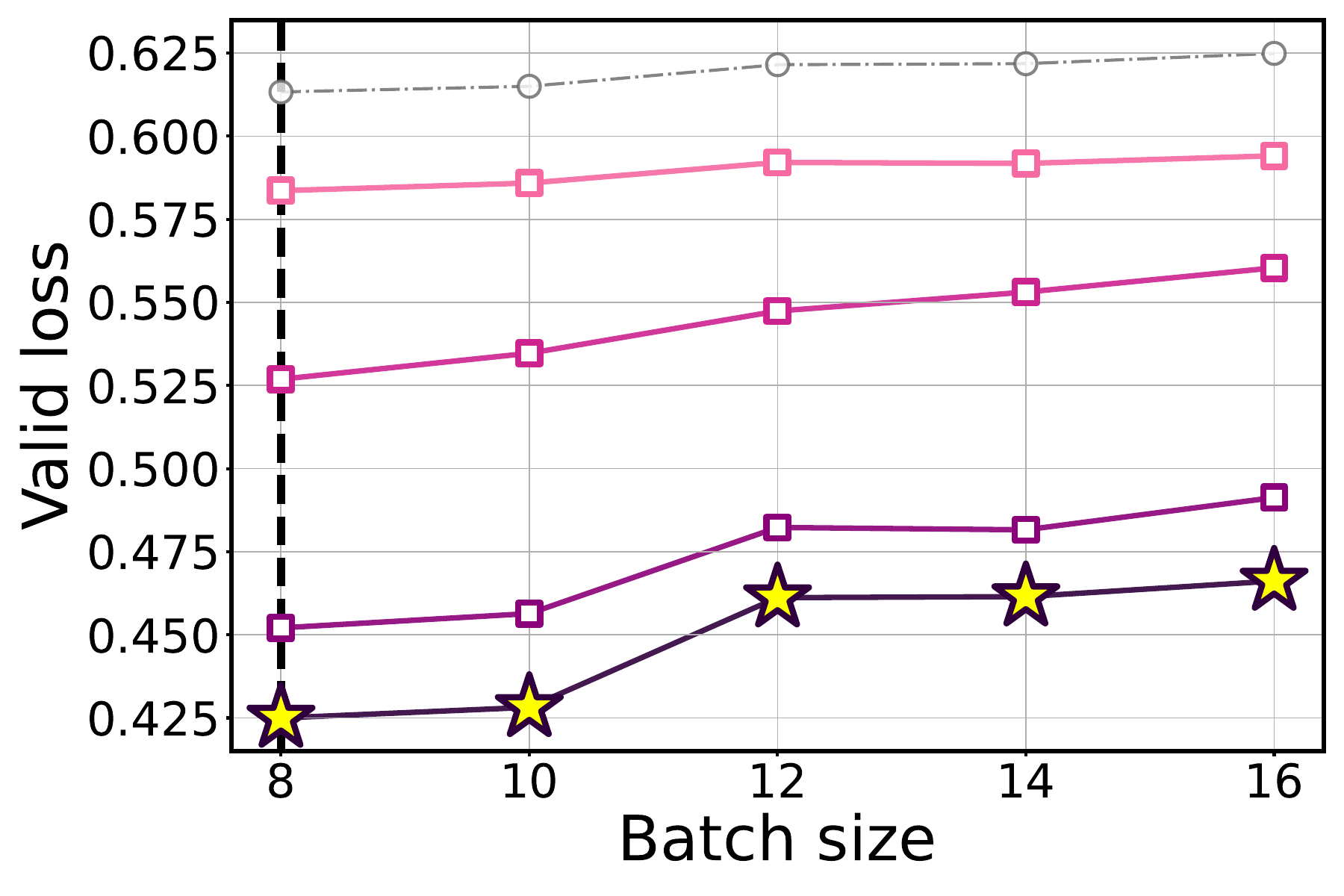}
    \includegraphics[width=0.48\textwidth]{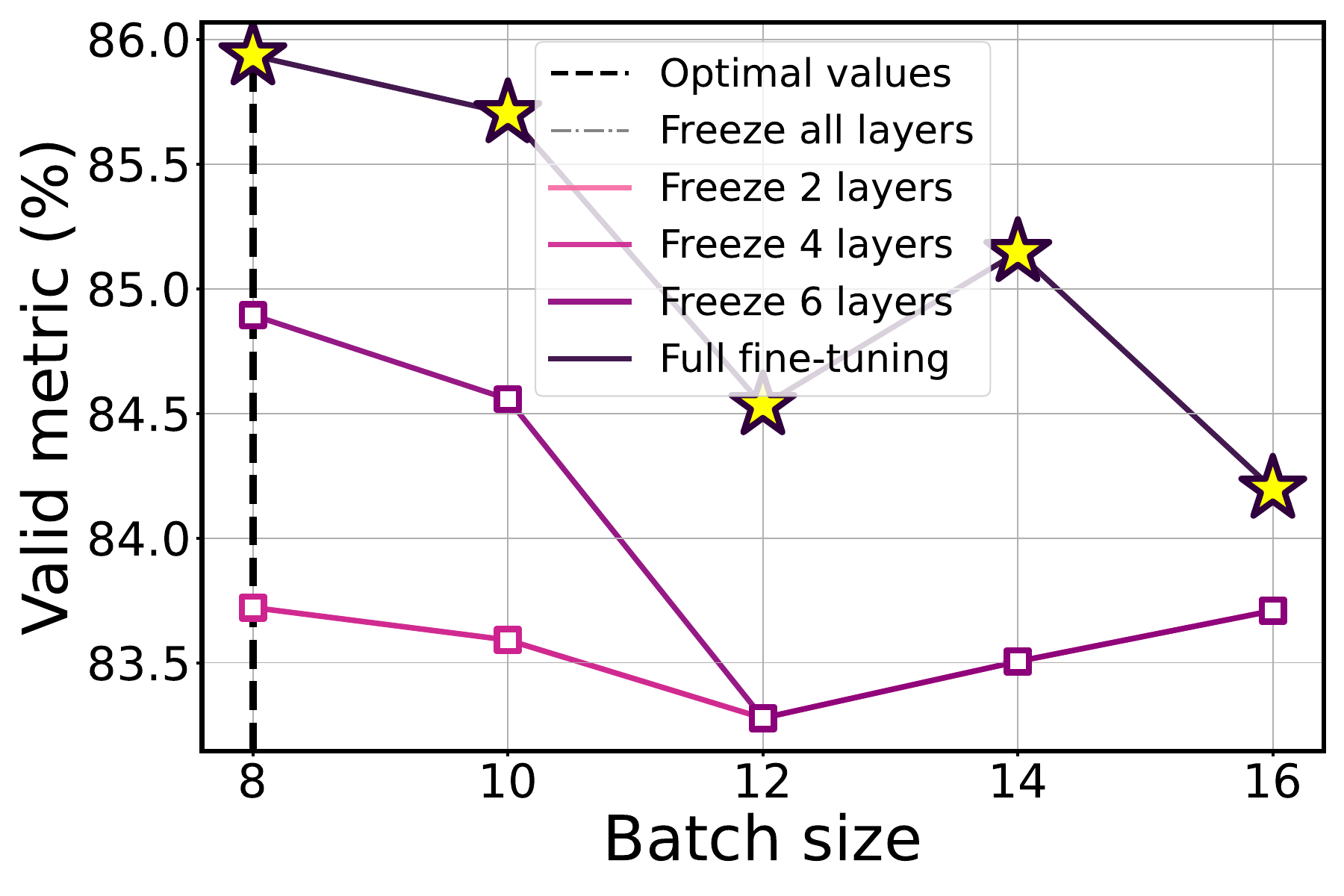}
    \caption{Validation results on the MRPC dataset for \gls{roberta}: loss (shown in left panels) and F1 score (in right panels) for varying learning rates, batch sizes, and frozen layers. Optimal hyperparameters align well across different frozen layers, except when all pre-trained layers are frozen.}
    \label{main:fig:hyperparameter_alignment}
\end{figure}

Discovering the optimal training hyperparameters incurs significant computational costs, particularly when fine-tuning extensive foundational models~\citep{aghajanyan2020better, mosbach2020stability, wang-etal-2023-two}. 
This challenge arises since the ideal set of hyperparameters tends to vary in tandem with changes in both tasks and model structures.

Surprisingly, our empirical findings reveal a consistent alignment of optimal hyperparameters when fine-tuning \glspl{plm}, regardless of variations in the number of frozen layers or the rank of \gls{lora}. As illustrated in \cref{main:fig:hyperparameter_alignment}, the alterations in validation loss and metric resulting from changes in the learning rate or batch size exhibit a similar pattern across different numbers of frozen layers, except in the case when all pre-trained layers are frozen and only the classifier layer is trained. This proves that we can decrease computational cost for searching the optimal hyperparameters by tuning on smaller models with more frozen layers or \gls{lora} with smaller ranks. Refer to \cref{app:hp_align} to see the additional results when varying the adam beta, learning rate schedule, as well as the case of the \gls{lora}. 

\citet{yang2022tensor} demonstrate that employing a particular model weight initialization method and learning rate scheduling method, referred to as $\mu$-parametrization, enables the transferability of certain training hyperparameters (such as learning rate and momentum) varying the width of the model. However, it is important to note that these results specifically pertain to scenarios where models are trained from scratch. This distinction is noteworthy as our context involves the fine-tuning of \glspl{plm}. It would be a great future research direction to theoretically analyze this phenomenon.
\section{Bayesian Optimization Model Fusion}
\label{main:sec:method}

In this section, \gls{ours} unfolds in three key steps. In \cref{main:subsec:fusion_members}, we present the process of constructing a set of fine-tuned trainable weights $\calS$, serving as components for model fusion. In \cref{main:subsec:hyperparam}, we introduce a method to identify optimal hyperparameters crucial in the construction of the set $\calS$ based on the findings explained in \cref{main:subsec:hyperparam_alignment}. Finally, we delve into how we conduct weighted averaging in \cref{main:subsec:MOBO}, following the insights discussed in \cref{main:subsec:landscape}.

\subsection{Fusion Member Sampling}
\label{main:subsec:fusion_members}

To improve the performance of our model through model fusion, it is crucial to carefully create the set $\calS$ by employing an appropriate weight sampling method. There are two main types of weight sampling methods: 1) sampling from multiple training trajectories~\citep{wortsman2022model} and 2) sampling from a single training trajectory with proper learning rate scheduling~\citep{izmailovaveraging}. However, \citet{wortsman2022model} indicate that, when applying model fusion with samples from multiple training trajectories, the performance improvement becomes less significant during the fine-tuning of \glspl{plm} compared to vision tasks. This limitation in \gls{nlp} tasks is attributed to the misalignment in loss and metric surfaces, as discussed in~\cref{main:subsec:landscape}.
Furthermore, when employing multiple training trajectories to sample fusion members, the training computation cost increases linearly in proportion to the number of fusion members. This poses a significant challenge, particularly in the context of fine-tuning \glspl{plm}. For these reasons, in our approach, we collect our fusion members from a single training trajectory. Since the fine-tuning process of \glspl{plm} involves a small number of training epochs and exhibits rapid convergence~\citep{lu2022improving}, we start gathering fusion members after 50\% of the training epochs are completed. This timing is slightly quicker than the point described in the work~\citep{izmailovaveraging},
which begins collecting after 75\% of the training epochs are concluded.
Once we start collecting the fusion members, we proceed to uniformly sample 15 members throughout the remaining training epochs.
Refer to~\cref{app:sec:experimental_details} for more details on the process of collecting fusion members.

\subsection{Hyperparameter Search via Bayesian Optimization}
\label{main:subsec:hyperparam}

\begin{wrapfigure}{R}{0.43\textwidth}
    \vspace{-15pt}
    \centering
    \includegraphics[width=0.34\textwidth]{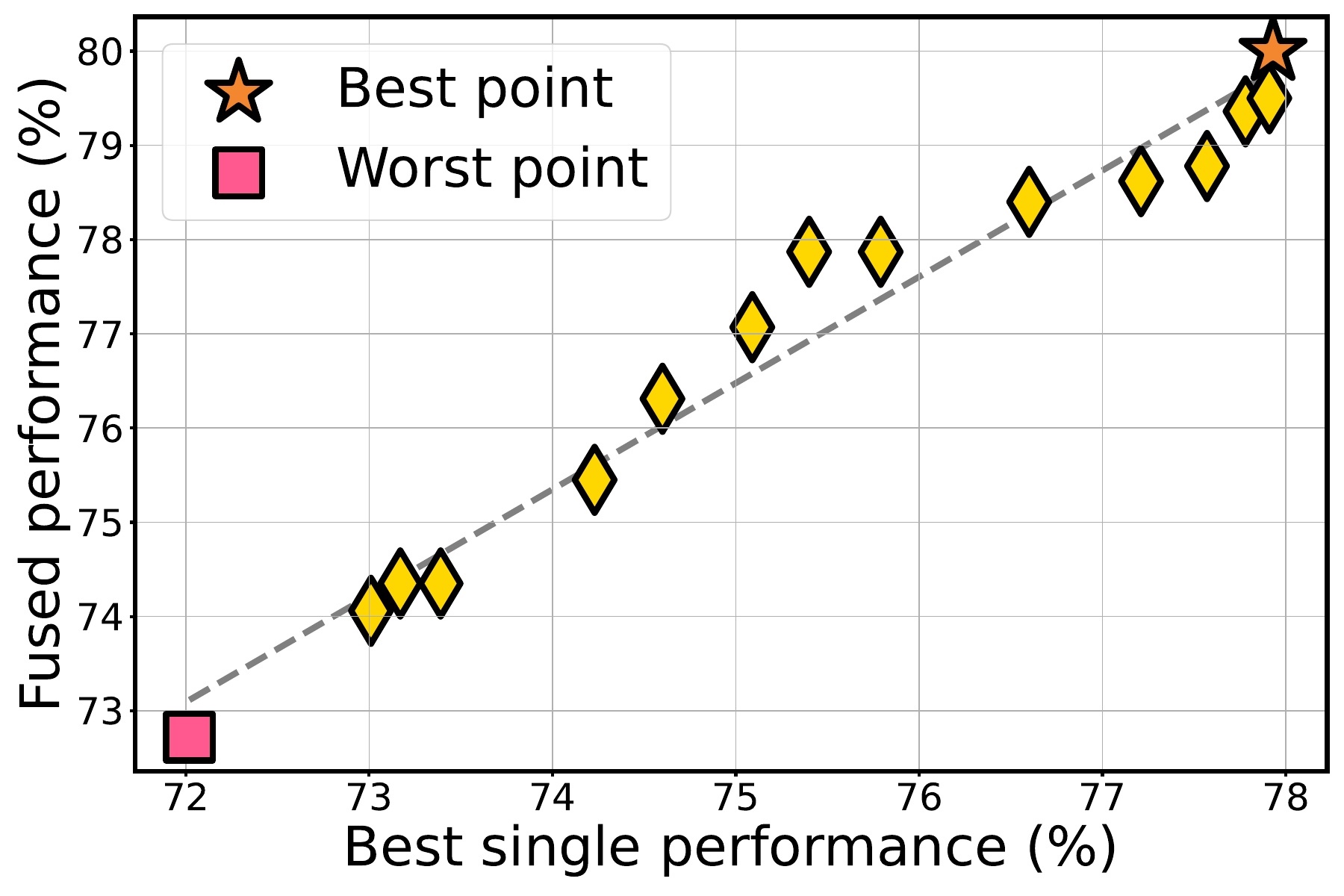}
    \caption{Correlation between the performance of best-performing weights in a training trajectory and the performance of the fused model. We fine-tune the \gls{roberta} model on the RTE dataset. Each point is obtained from the evaluation of a single trajectory with varying hyperparameters.}
    \label{main:fig:correlation}
    \vspace{-10pt}
\end{wrapfigure}

In the construction of a set of fusion members $\calS$ from a single training trajectory, the effectiveness of the training trajectory significantly impacts the ultimate metric performance of the fused model weight $\Bar{\bw}$. In this context, the effectiveness of a training trajectory refers to the model's metric performance using the best-performing weight within that trajectory on the validation dataset $\calD_{\text{val}}$. The correlation in \cref{main:fig:correlation} strongly indicates that the performance of the best-performing weight is positively correlated with the performance of the fused weight. Consequently, to achieve the best performance of the fused weight, it becomes crucial to identify the set of optimal hyperparameters $\blambda$ that results in the most effective training trajectory. However, two primary challenges arise when searching for the optimal hyperparameters $\blambda^\star$ that yield the best metric performance: 1) the metric functions $\{\fmetric^{(k)}\}_{k=1}^K$ are non-differentiable and 2) we need to efficiently assign computational resources in finding better hyperparameters beyond na\"ive methods such as grid search.
To remedy these two issues simultaneously, in \gls{ours}, we employ \gls{bo} to find the optimal set of hyperparameters:
\begin{align}
    \blambda^* = \argmin_{\lambda} \sum_{k=1}^K \Bar{f}_{\text{metric}}^{(k)}(\calM(\btheta_{\text{init}},\bw(\blambda)), \calD_{\text{val}}),
\end{align}
where $\bw(\blambda)$ represents the best-performing weight within the training trajectory associated with the hyperparameter set $\blambda$. Here, we utilize \gls{gp} regression~\citep{RasmussenCE2006book} and Log-Expected Improvement~\citep{ament2023unexpected} as a surrogate function and an acquisition function, respectively. We employ three randomly initialized sets of hyperparameters as the starting point for \gls{bo}, conducting 10 iterations of computations to determine the optimal set $\blambda^\star$.

The sequential nature of \gls{bo} computations can lead to a substantial computational load, particularly in the context of fine-tuning \glspl{plm}. To address this issue and propose a more computationally efficient \gls{bo} approach, we draw insights from the observations discussed in \cref{main:subsec:hyperparam_alignment}. The alignment of the best hyperparameters for fine-tuning between the full model and lightweight models (e.g., frozen layers model or reduced rank \gls{lora}) allows us to utilize the lightweight model instead of the full model when seeking the optimal set $\blambda^\star$ as follows:
\begin{align}
    \blambda^\star = \argmin_{\lambda} \sum_{k=1}^K \Bar{f}_{\text{metric}}^{(k)}(\calM(\btheta_{\text{init}}, \widehat{\bw}(\blambda)), \calD_{\text{val}}),
\end{align}
where $\widehat{\bw}$ is the trainable weight of the lightweight model. Refer to \cref{main:sec:experiments} to see how our computationally efficient method decreases computation time while maintaining performance.

\subsection{Multi-Objective Bayesian Optimization for Model Fusion}
\label{main:subsec:MOBO}

After completing the construction of the set $\calS$ with $N$ individual models, the next stage involves selecting appropriate averaging coefficients $\bdelta \in [0, 1]^N$ to ensure the enhanced metric performance of a fused model. To achieve this, we can leverage metrics $\{\fmetric^{(k)}\}_{k=1}^K$ and apply a \gls{bo} procedure to obtain optimal averaging coefficients $\bdelta^\star$, similar to the optimization process for the hyperparameter set $\blambda$. However, restricting the consideration to metric performance solely on $\calD_{\text{val}}$ may result in our fused weights $\Bar{\bw}$ overfitting to $\calD_{\text{val}}$ and exhibiting poor generalization to the true data distribution, due to the complex and sharp nature of the metric landscape which is observed in \cref{main:subsec:landscape}. To tackle this challenge, when optimizing $\bdelta$, we propose to minimize both $\floss$ and $\{\fmetric^{(k)}\}_{k=1}^K$ by employing \gls{mobo} identify a Pareto frontier defined as follows:
\begin{align}
    \calP = \Big\{\bdelta^\star \mid \bdelta^\star = \argmin_{\bdelta} \big( l(\bdelta), l_1(\bdelta), \ldots, l_{K}(\bdelta)\big)\Big\},
\end{align}
where $l(\bdelta) \coloneqq \Bar{f}_{\text{loss}}(\calM(\btheta_{\text{init}},\Bar{\bw}(\bdelta)), \calD_{\text{val}})$ and $l_k(\bdelta) \coloneqq \Bar{f}_{\text{metric}}^{(k)}(\calM(\btheta_{\text{init}},\Bar{\bw}(\bdelta)), \calD_{\text{val}})$ for $k\in[K]$.
Note that $\Bar{\bw}(\bdelta)$ denotes a fused set of weights with an averaging coefficient vector $\bdelta$, i.e., $\Bar{\bw}(\bdelta)=\sum_{i=1}^N\delta_i\bw_i$ where $\bw_i\in\calS$ for $i\in[N]$
and $N$ is the number of models to fuse.

Here, we utilize the \gls{ehvi} strategy, which is described in the work by~\citet{EmmerichMTM2006ieeetec}.
The hypervolume, in this context, is defined as a volume size between $\calP$ and a reference point $\br$.
We set the reference points as a zero vector.
To enhance the optimization of the hypervolume improvement objective, we employ the logarithmic form of $q$NEHVI algorithm~\citep{DaultonS2021neurips,ament2023unexpected},
which is implemented with the BoTorch framework~\citep{balandat2020botorch}.
As highlighted in~\cref{main:sec:back}, this algorithm has proven effective in practical multi-objective optimization scenarios. This makes it well-suited to handle the complex and sharp nature of our metric landscape, enabling it to successfully identify the optimal $\bdelta^\star$.
We run \gls{mobo} for a total of $5|\bdelta|=75$ iterations to find the optimal coefficients $\bdelta^\star$.

In our case, additional constraints are in place for executing \gls{mobo}, specifically 1) equality constraints and 2) inequality constraints for $\bdelta$. To address the inequality constraints (i.e., $\delta_i \geq 0$), we follow the work by~\citet{gardner2014bayesian} to incorporate constraints into the acquisition function. To deal with the equality constraints $\sum_{i=1}^{N} \delta_i = 1$, we simply normalize the output of the acquisition function. Refer to~\cref{algorithm/ours} in~\cref{app:sec:algorithm} for the summary of \gls{ours}.

\section{Related Work}




\paragraph{Model Fusion for Pre-Trained Language Models.}
Due to the increasing number of model parameters in recent \glspl{plm}, there has been a significant increase in both memory requirements and computational costs~\citep{zhang2022opt,chowdhery2023palm,touvron2023llama}. Consequently, there is growing attention on a research direction aimed at enhancing the performance of \glspl{plm} while simultaneously managing computational costs and memory requirements through the exploration of model fusion methods~\citep{rame2023rewardedsoups,yadav2023resolving,chronopoulou2023adaptersoup}. 
However, most of these studies have focused on fusing the models fine-tuned on different tasks, aiming to develop a single multi-task learner. In the context of a single-task fine-tuning scenario within \gls{plm}, it has been observed that the previous simple weight-averaging approaches often yield marginal improvements~\citep{wortsman2022model,kaddour2022flat}; nevertheless, the exploration into the underlying rationale of this consequence remains limited. As mentioned in~\cref{main:sec:mismatch}, we find that uniform weight averaging does not always align generalization on the loss surface with the optimal point on the metric surface, primarily due to the discrepancy between loss and metric landscapes. To address this issue, we develope a single-task model fusion method based on \gls{mobo}, finding the optimal weight combination coefficients by considering both metrics and loss functions.

\paragraph{Bayesian Optimization.}

\gls{bo}~\citep{BrochuE2010arxiv,GarnettR2023book} is a promising strategy to optimize a black-box function.
In particular, if a target objective is costly in terms of function evaluations,
Specifically, \gls{bo} sequentially seeks solution candidates by modeling a surrogate function and maximizing an acquisition function.
In the \gls{bo} community, a \gls{gp}~\citep{RasmussenCE2006book} is often employed as a surrogate function but diverse regression models such as Bayesian neural networks~\citep{SpringenbergJT2016neurips,LiYL2023arxiv} and tree-based models~\citep{HutterF2011lion,KimJ2022aistats} can be used.
As a choice of acquisition function,
expected improvement~\citep{JonesDR1998jgo} and \gls{gp} upper confidence bound~\citep{SrinivasN2010icml} are often considered.
Importantly,
\gls{bo} is more effective than other existing optimization strategies such as grid search and genetic algorithms~\citep{TurnerR2020neuripscd}.
Its efficacy has been demonstrated in a wide variety of applications such as hyperparameter optimization~\citep{SnoekJ2012neurips}, nanostructured device design~\citep{KimJ2024dd}, and chemical reaction optimization~\citep{ShieldsBJ2021nature}.
Moreover, in the deep learning context, the necessity for efficient hyperparameter tuning via \gls{bo} has risen following the increasing number of hyperparameters and parameters in models~\citep{snoek2015scalable}. Consequently, \gls{bo} is applied for hyperparameter optimization in various deep learning tasks, such as image classification~\citep{KandasamyK2018neurips,LethamB2020neurips} and \gls{nlp} tasks~\citep{melis2017state,chen2023instructzero}.
\section{Experiments}
\label{main:sec:experiments}

\begin{table*}[t]
\centering
\caption{\textbf{Results on Medium-Sized Language Models.} We conduct the text classification task using \gls{robertabase} on a subset of the GLUE benchmark datasets, and the question-answering task using \gls{t5}-base on the SQuAD2.0 dataset. ACC, F1, and EM denote accuracy, F1 score, and Exact Match, respectively.}
\label{tab:main_exp1}
\begin{scriptsize}
\begin{sc}
\setlength{\tabcolsep}{4pt}
    \begin{tabular}{lcccccccc}
\toprule
    & \multicolumn{6}{c}{\gls{robertabase}} & \multicolumn{1}{c}{\gls{t5}-base} \\
    \cmidrule(lr){2-7} \cmidrule(lr){8-8}
    Method & RTE (Acc) & MRPC (F1) & SST-2 (Acc) & QNLI (Acc) & QQP (F1) & MNLI (Acc) & SQuAD2.0 (F1/EM) \\
\midrule
Grid Fine-Tune & 77.78 & 92.39 & 94.87 & 92.62  & 88.16 & 87.41 & 78.18/72.83\\
\midrule
HPBO (Full) & 78.57 & 92.78 & 95.11 & 93.01  & 88.58 & 87.46 & 78.28/73.29\\
SWA & 78.62 & 92.24 & 95.42 & 92.81  & 88.49 & 87.41 & 80.31/74.85 \\
OTfusion & 77.08 & 92.82 & 94.27 & 92.22 & 88.34 & 87.43 & 80.75/74.99 \\
Greedy SWA & 80.70 & 92.83 & \UL{95.54} & 93.16  & 88.64 & 87.45 & 80.63/75.44\\
Learned SWA & \UL{81.40} & 92.81 & 95.31 & 92.94  & 88.38 & 87.41 & 80.65/74.23\\
TWA & 81.23 & 91.58 & \UL{95.54} & 93.00 & 87.85 & 87.42 & 80.29/74.79 \\
\midrule
$\text{BOMF$^{\dagger}$ (ours)}$ & \BL{81.75} & \UL{93.37} & \BL{95.65} & \BL{94.83} & \UL{88.66} & \UL{87.51} & \UL{80.82}/\UL{75.79}\\
BOMF (ours) & \UL{81.40} & \BL{93.90} & \UL{95.54} & \UL{93.50}  & \BL{88.68} & \BL{87.86} & \BL{81.82}/\BL{76.21}\\

\bottomrule
\end{tabular}
\end{sc}
\end{scriptsize}
\end{table*}

In this section, we present empirical results demonstrating the effectiveness of \gls{ours} in various \gls{nlp} tasks. We compare our method to five basic algorithms aimed at finding a high-performing solution. \textbf{Grid Fine-Tune} is a simple fine-tuning method that selects the best-performing checkpoint using grid search. \textbf{HPBO} utilizes optimal hyperparameters obtained by \cref{main:subsec:hyperparam} for fine-tuning the baselines. \textbf{SWA} is an optimization technique that averages model parameters obtained during training. \textbf{Greedy SWA} is a modified version of \gls{swa} inspired by Greedy Soups~\citep{wortsman2022model}, sorting weights based on metric performance on $\calD_{\text{val}}$ and including them in $\Bar{\calS}$ only if they improve $\Bar{\bw}$'s performance. \textbf{Learned SWA}, inspired by Learned Soup~\citep{wortsman2022model}, learns the coefficients $\bdelta$ based on the loss after fine-tuning. For medium-sized language models, we tested a variant of Transformer OTfusion~\citep{imfeld2023transformer}, aligning pre-trained weights before averaging. Additionally, we experimented with \textbf{TWA}~\citep{li2023trainable}, a recent \gls{swa} variant that reconstructs $\calS$ by finding weight space basis vectors and learns $\bdelta$ based on the loss.

In all tables, the best performance is indicated with \BL{boldfaced underline}, while the second-best value is represented with \UL{underline} in each column. The final column `Avg.' provides a summary of overall results for each method across various datasets or metrics. The terms `Full' and `Freeze' in~\cref{tab:main_exp1} specify the exploration of optimal hyperparameters using either the entire model or a model with half of its weights frozen, as discussed in~\cref{main:subsec:hyperparam}. Similarly, the terms `Rank 64' and `Rank 4' in~\cref{tab:large_exp1} denote that we use the Rank 64 or the lightweight Rank 4 version of the \gls{lora} model for the hyperparameter search,
respectively.
See~\cref{app:sec:experimental_details} for the details of downstream datasets and hyperparameter selection.

\subsection{Empirical Analysis on Medium-Sized Language Models}

We begin by evaluating the effectiveness of \gls{ours} on medium-sized language models using \gls{robertabase}~\citep{liu2019} and \gls{t5}-base~\citep{raffel2020exploring}. For \gls{robertabase}, we performed text classification tasks using the GLUE benchmark datasets~\citep{wang2018glue}. For \gls{t5}-base, we carried out the question-answering task with the SQuAD2.0~\citep{rajpurkar2018know} dataset. For both models, we fine-tuned the weights directly on the downstream datasets.

\cref{tab:main_exp1} shows that \gls{ours} consistently outperforms other baselines across all model structure and datasets.\footnote{A $\dagger$ symbol indicates results from the trajectory found using full HPBO, while results without the symbol indicate trajectories found using freeze HPBO.}
Notably, the performance of HPBO, which uses hyperparameters obtained from \cref{main:subsec:hyperparam} with the full model, surpasses Grid Fine-Tune for most datasets. These results demonstrate that our \gls{bo}-based hyperparameter search framework effectively discovers optimal hyperparameters compared to grid search. Refer to \cref{app:full_exps} for the performance of freeze HPBO, which uses a lightweight model for hyperparameter optimization. Freeze HPBO also clearly outperforms Grid Fine-Tune which proves the effectiveness of our \gls{bo}-based hyperparameter search. Also, it is evident that model fusion methods, except \gls{ours}, lead to performance declines compared to HPBO, as discussed in~\cref{main:subsec:landscape}, in certain datasets. On the contrary, \gls{ours} consistently betters the performance compared to HPBO, yielding that our method with \gls{mobo} effectively finds optimal $\bdelta^\star$ even in complex and sharp metric landscapes. Refer to \cref{tab:large_exp1_full} for the complete results.

\subsection{Empirical Analysis on Large Language Models}

\begin{table}[t]
    \centering
    \caption{\textbf{Results on Large Language Models.} We compare the results of \gls{ours} and baseline methods using the SAMSum and KorMedMCQA datasets for summarization and medical multiple choice question-answering tasks with \gls{llama27b} and \gls{llama38b}. R1, R2, and RL denote Rouge-1, Rouge-2, and Rouge-L scores for summarization. Doctor, Nurse, and Pharm denote evaluation results for medical question answering in each respective field, using accuracy as the metric.}
    \label{tab:combined_results}
    \begin{minipage}[t]{0.48\textwidth}
        \centering
        \subcaption{Summarization (SAMSum)}
        \label{tab:large_exp1}
        \scriptsize
        \begin{sc}
        \setlength{\tabcolsep}{5pt}
        \begin{tabular}{lcccc}
        \toprule
        Method & R1 & R2 & RL & Avg. \\ 
        \midrule
        \addlinespace[2pt]
        HPBO (Rank 64) & 52.66 & 28.22 & 44.33 & 41.73\\
        \addlinespace[2pt]
        SWA & 51.81 & 27.61 & 43.55 & 40.99 \\
        \addlinespace[2pt]
        Greedy SWA & \BL{53.40} & 28.06 & 43.31 & 41.49 \\
        \addlinespace[2pt]
        Learned SWA & 52.93 & \BL{28.97} & 44.04 & 41.98 \\ 
        \addlinespace[1pt]
        \midrule
        BOMF$^{\dagger}$ (ours) & \BL{53.40} & \UL{28.78} & \UL{44.38} & \BL{42.19} \\ 
        \addlinespace[2pt]
        BOMF (ours) & \UL{53.07} & 28.61 & \BL{44.40} & \UL{42.03}  \\
        \bottomrule
        \end{tabular}
        \end{sc}
    \end{minipage}
    \hfill
    \begin{minipage}[t]{0.48\textwidth}
        \centering
        \subcaption{Korean Medical Question Answering}
        \label{tab:large_exp2}
        \scriptsize
        \begin{sc}
        \setlength{\tabcolsep}{4pt}
        \begin{tabular}{lccccc}
        \toprule
        Method & Doctor & Nurse & Pharm & Avg.\\
        \midrule
        ICL & 37.89 & 50.15 & 50.00 & 46.01 \\
        \midrule
        HPBO (Rank 64) & 43.62 & 54.64 & 51.49 &  49.92 \\
        SWA & 43.96 & 54.64 & 51.97 & 50.19 \\
        Greedy SWA & 43.97 & 54.64 & 51.98 & 50.20\\
        Learned SWA & 44.06 & 54.94 & 52.28 & 50.43 \\
        \midrule
        BOMF$^{\dagger}$ (ours) & \UL{45.00} & \BL{55.70} & \BL{52.97} & \BL{51.22} \\
        BOMF (ours) & \BL{45.31} & \UL{55.37} & \UL{52.80} & \UL{51.16} \\
        \bottomrule
        \end{tabular}
        \end{sc}
    \end{minipage}
\end{table}

We further validated the effectiveness of our proposed method by fine-tuning larger models using \gls{lora}. Specifically, we experimented with \gls{llama27b} and \gls{llama38b} on tasks such as summarization using the SAMsum~\citep{gliwa2019samsum} dataset, Korean multi-choice medical question answering using the KorMedMCQA~\citep{kweon2024kormedmcqa} dataset, and dialogue generation using the E2E~\citep{novikova2017e2e} dataset. In the summarization task, while Learned SWA exhibited the best performance in terms of Rouge-2, \gls{ours} surpassed Learned SWA in average performance across all metrics, as illustrated in \cref{tab:large_exp1}. Notably, for Rouge-L, only \gls{ours} improved over HPBO, highlighting the effectiveness of the multi-objective approach in \gls{ours}. Furthermore, as shown in \cref{tab:large_exp2}, our model not only outperforms other baselines but also demonstrates that fine-tuning remains essential for specific tasks despite the rise of in-context learning (ICL)~\citep{dong2022icl}. This highlights the necessity of \gls{ours}, which efficiently identifies hyperparameters and provides an effective fine-tuning solution through model fusion. The results for E2E can be found in \cref{app:full_exps}.

\subsection{Ablation Study}

\begin{wraptable}{r}{0.48\textwidth}
\vspace{-15pt}
\centering
\caption{\textbf{Results on the Varying Number of Frozen Layers.} Comparison of the number of parameters and relative training wall-clock time per epoch when optimizing hyperparameters across different numbers of frozen layers, using \gls{robertabase} fine-tuned on the RTE and MRPC datasets.}
\label{tab:ab:num_layers}
\scriptsize
\begin{sc}
\setlength{\tabcolsep}{2pt}
    \begin{tabular}{lcccc}
\toprule
Task & Params &  Relative time & RTE & MRPC \\ 
\midrule
Grid Fine-Tune & 125m & $\times$1 & 77.78 & 92.39\\
\midrule
Full & 125m & $\times$1  & 78.57 & 92.78 \\ 
Freeze 2 & 110m & $\times$0.53 & 78.50 & 92.39 \\ 
Freeze 4 & 96m & $\times$0.44 & 78.34 & 92.36 \\ 
Freeze 6 & 82m & $\times$0.34 & 78.49 & 92.72 \\ 
\bottomrule
\end{tabular}
\end{sc}
\vspace{-10pt}
\end{wraptable}

\paragraph{Number of Frozen Layers.}
To analyze the efficiency of memory and compute when using a lightweight model in the \gls{bo} procedure to find $\blambda^\star$, we conduct a study using \gls{robertabase} on the RTE and MRPC datasets. As presented in \cref{tab:ab:num_layers}, the use of a lightweight model successfully identifies favorable hyperparameters that yield good performance while reducing the number of parameters by up to 25\% and the computation time by up to 66\%. This efficiency is achieved by caching outputs from the frozen layers. By systematically freezing layers from the tail of the model, we can cache the outputs from these frozen layers and reuse them during the training process.

\paragraph{Multiple Objectives.}

\begin{wraptable}{r}{0.48\textwidth}
\vspace{-15pt}
\centering
\caption{\textbf{Comparison of Using Multi-Objective and Single-Objective Approaches.} Results of \gls{ours} and single-objective \gls{bo} baselines with \gls{t5}-base fine-tuned on the SQuAD2.0 dataset.}
\label{tab:ab:single-multi}
\scriptsize
\begin{sc}
\setlength{\tabcolsep}{12pt}
\begin{tabular}{lccc}
\toprule
Metric & F1 & EM & Avg. \\
\midrule
F1 only  & 81.01 & 75.09 & 78.05 \\
EM only  & 80.40 & 75.87 & 78.13 \\
BOMF & 80.82 & 75.79 & 78.31 \\
\bottomrule
\end{tabular}
\end{sc}
\vspace{-10pt}
\end{wraptable}

To validate the efficacy of using multiple objectives when determining optimal $\bdelta$, we compare \gls{ours} with single-objective baselines using \gls{t5}-base on the SQuAD2.0 dataset. In this task, we consider two metrics: F1 score and Exact Match. \cref{tab:ab:single-multi} shows that relying on only one specific metric slightly increases the objective metric but results in a significant performance drop for the other metric. This outcome suggests that using single-objective \gls{bo} is appropriate when aiming to find a model optimized for a specific metric, while the use of \gls{mobo} is more suitable for discovering an optimal fused model that achieves high performance across various metrics. Refer to~\cref{app:ablation} for further ablation studies.

\section{Conclusion}
\label{main:sec:conclusion}
In this paper, we empirically remarked two intriguing findings on loss and metric landscapes and hyperparameter alignment.
Then, motivated by the observations mentioned above, we proposed a novel \gls{bo}-based \gls{ours} algorithm for model fusion.
Our method utilizes the \gls{bo} and \gls{mobo} frameworks to seek optimal fine-tuning hyperparameters and averaging coefficients, respectively.
We validated that our proposed method exhibits improved performance on both \gls{nlu} and \gls{nlg} tasks on middle- and large-scale \glspl{plm}.

\paragraph{Limitations and Future Work.}
As discussed in \cref{main:subsec:hyperparam_alignment}, compelling future research involves the theoretical analysis of the hyperparameter alignment phenomenon. Moreover, we empirically observed that when utilizing quantization-based low-rank approximation methods~\citep{qlora2023dettmers,li2023loftq}, traditional uniform averaging methods and weighted averaging methods face challenges in effectively aggregating models. These challenges arise from the quantized weight values in the models that behave differently with averaging weights. Another research direction is the development of averaging methods for the quantization-based low-rank approximation methods.


\begin{ack}
This work was partly supported by the National Research Foundation of Korea (NRF) grant funded by the Korea government (MSIT) (NRF-2022R1A5A708390812) and Institute of Information $\&$ communications Technology Planning $\&$ Evaluation (IITP) grant funded by the Korea government (MSIT) (No.RS-2019-II190075, Artificial Intelligence Graduate School Program (KAIST), No.2022-0-00184, Development and Study of AI Technologies to Inexpensively Conform to Evolving Policy on Ethics, No.2022-0-00713, Meta-learning Applicable to Real-world Problems).
\end{ack}

\bibliography{references}
\bibliographystyle{abbrvnat}

\newpage
\appendix

\section{Details of Experiments}
\label{app:sec:experimental_details}

Our implementation leverages key libraries, including PyTorch 2.0.1~\citep{paszke2019pytorch},
Huggingface Transformers~\citep{wolf2019huggingface},
and BoTorch~\citep{balandat2020botorch},
to construct a robust framework for our experiments.
These experiments are rigorously conducted on high-performance computing hardware, specifically NVIDIA RTX 3090 and NVIDIA RTX A6000 GPUs, to ensure the efficiency and scalability of our models. To further bolster the reproducibility of our results, we meticulously set and documented all experiment seeds, enabling precise replication of our experimental conditions and findings.


\subsection{Medium-Sized Language Models}

\begin{table}[ht]
\centering
\caption{Detailed \gls{roberta} experimental setup.}
\label{roberta-exp-details}
\renewcommand{\arraystretch}{1.5}
\begin{center}
\begin{sc}
\begin{tabular}{ll}
\toprule
\textbf{Category} & \textbf{Details} \\
\midrule
\multicolumn{2}{c}{\textbf{Model Specifications}} \\
\midrule
Architecture & Transformer \\
Pre-training & \gls{roberta}-base \\
Optimizer & AdamW \\
Scheduler & Linear scheduler with warmup\\
Warmup ratio & 0.2 If RTE or MRPC else 0.1\\
Learning rate & [1e-06, 1e-04] \\
Batch size & [8, 16] If RTE or MRPC else [32, 64] \\
Epochs & 20 if RTE else 10 \\
\midrule
\multicolumn{2}{c}{\textbf{Task Specifications}} \\
\midrule
\textbf{Task name} & Classification \\
Dataset & Subset of GLUE benchmarks.\\
\bottomrule
\end{tabular}
\end{sc}
\end{center}
\end{table}

For the \gls{roberta} model, we evaluated the performance for classification and utilized a subset of the GLUE benchmark~\citep{wang2018glue}. This benchmark serves as a comprehensive evaluation of a language model's overall \gls{nlu} capabilities. The Recognizing Textual Entailment (RTE) task, which employs neutral and contradiction instances to assign a not-entailment label, is a binary classification task comprising 2,490 training instances and 277 validation instances. The Microsoft Research Paraphrase Corpus (MRPC)~\citep{dolan2005automatically} consists of sentence pairs and corresponding labels. This task involves binary classification to determine whether a pair of sentences are semantically equivalent, utilizing the F1 score as the metric due to label imbalance. This dataset contains a total of 3,668 training and 408 validation instances. The Stanford Sentiment Treebank (SST-2)~\citep{socher2013recursive} includes movie reviews with associated positive/negative labels. The task is binary classification to discern the sentiment of a given sentence as positive or negative, with 67,349 training and 872 validation instances. The Stanford Question Answering Dataset (QNLI)~\citep{rajpurkar2018know} is a question-answering task composed of paragraph-question pairs, where one sentence in the paragraph contains the answer to the human-generated question. This dataset comprises 104,743 training and 5,463 validation instances. The Quora Question Pairs dataset (QQP)~\citep{wang2017bilateral} involves determining whether two questions are semantically equivalent, again using the F1 score as the metric due to label imbalance, with 363,846 training and 40,430 test instances. Lastly, The Multi-Genre Natural Language Inference Corpus (MNLI)~\citep{williams2017broad} is labeled for textual entailment across genre pairs, primarily consisting of premise and hypothesis sentence pairs. This task predicts the relationship between these sentences in three categories. The dataset includes 392,702 training and 9,815 validation instances, of which we used the matched case of the validation set. We conducted experiments by selecting two datasets from each GLUE benchmark based on their size scale. Additionally, specific details on the fine-tuning methods can be found in \cref{roberta-exp-details}.

\begin{table}[ht]
\caption{Detailed T5 experimental setup.}
\label{t5-exp-details}
\renewcommand{\arraystretch}{1.5}
\begin{center}
\setlength{\tabcolsep}{20pt}
\begin{tabular}{ll}
\toprule
\textbf{\textsc{Category}} & \textbf{\textsc{Details}} \\
\midrule
\multicolumn{2}{c}{\textbf{\textsc{Model Specifications}}} \\
\midrule
\textsc{Architecture} & \textsc{Transformer} \\
\textsc{Pre-training} & \textsc{\gls{t5}-base} \\
\textsc{Optimizer} & \textsc{AdamW} \\
\textsc{Learning rate} & \textsc{[1e-06, 1e-04]} \\
\textsc{Batch size} & \textsc{[32, 64]} \\
\textsc{Gradient accumulation step} & \textsc{2} \\
\textsc{Epochs} & \textsc{3.0} \\
\midrule
\multicolumn{2}{c}{\textbf{\textsc{Task Specifications}}} \\
\midrule
\textbf{\textsc{Task name}} & Question Answering \\
\textsc{Input text} & \textit{``question: \{question\} context: \{context\}''} \\
\textsc{Label text} & \textit{``\{answer\}''} \\
\textsc{Dataset} & \textsc{SQuAD2.0} \\
\midrule
\textsc{Max new tokens} & \textsc{10} \\
\bottomrule
\end{tabular}
\end{center}
\end{table}

For the \gls{t5}-base model, we utilize the Stanford Question Answering Dataset (SQuAD 2.0)~\citep{rajpurkar2018know}. This dataset comprises 130,319 training pairs and 11,873 validation pairs of questions and answers. The dataset can be accessed through the Hugging Face datasets library.\footnote{\url{https://huggingface.co/datasets/squad_v2}} Details on our fine-tuning procedures are provided in \cref{t5-exp-details}. Furthermore, we assess the generated answers by adhering to the code established in the official SQuAD 2.0 repository.\footnote{\url{https://rajpurkar.github.io/SQuAD-explorer}}

\subsection{Large Language Models}

In our experiments with the \gls{llama27b}\footnote{\url{https://huggingface.co/meta-llama/Llama-2-7b-hf}} model, we focused on two tasks: summarization and dialogue generation. For the summarization task, we employed the Samsung Abstractive Messenger Summarization (SAMSum) dataset~\citep{gliwa2019samsum}, which consists of 14,732 training samples, 818 validation samples, and 819 test samples. For the dialogue generation task, we selected the End-to-End NLG Challenge (E2E) dataset~\citep{novikova2017e2e}. This dataset includes 42,061 training samples, 4,672 validation samples, and 4,693 test samples. Details of our fine-tuning process are provided in \cref{llama-exp-details}. Notably, in the case of the E2E dataset, the test set typically contains around five common inputs with a variety of labels. To save time, we conducted a generate process for one common input and used the different labels as multiple references to calculate the metrics. Consequently, for evaluation, the sentences generated by the model are based on a unique label, totaling 630 sentences. This accounts for the discrepancy in experimental performance between our study and that presented in the original paper of the E2E dataset~\citep{novikova2017e2e}. All metrics including BLEU, METEOR, and ROUGE were computed using the Huggingface evaluate library.\footnote{\url{https://huggingface.co/evaluate}}

\begin{table}[ht]
\caption{Detailed \gls{llama27b} experimental setup.}
\label{llama-exp-details}
\renewcommand{\arraystretch}{1.5}
\begin{center}
\setlength{\tabcolsep}{2pt}
\begin{tabular}{ll}
\toprule
\textbf{\textsc{Category}} & \textbf{\textsc{Details}} \\
\midrule
\multicolumn{2}{c}{\textbf{\textsc{Model Specifications}}} \\
\midrule
\textsc{Architecture} & \textsc{Transformer} \\
\textsc{Pre-training} & \textsc{LLaMA-2-7B} \\
\textsc{LoRA alpha} & \textsc{16} \\
\textsc{LoRA dropout} & \textsc{0.1} \\
\textsc{Optimizer} & \textsc{AdamW} \\
\textsc{Learning rate} & \textsc{[1e-06, 1e-03]} \\
\textsc{Batch size} & \textsc{[16, 32]} \\
\textsc{Gradient accumulation step} & \textsc{2} \\
\textsc{Epochs} & \textsc{2} \\
\midrule
\multicolumn{2}{c}{\textbf{\textsc{Task Specifications}}} \\
\midrule
\textbf{\textsc{Task name}} & \textsc{Summarization} \\
\textsc{Prompt} & \textit{"Summarize the following dialogue that is delimited}\\
& \textit{with triple backticks."} \\
\textsc{Dataset} & \textsc{SamSum} \\
\midrule
\textbf{\textsc{Task name}} & \textsc{Dialogue Generation} \\
\textsc{Prompt} & \textit{"Generate a natural language description for the following}\\
& \textit{restaurant attributes."} \\
\textsc{Dataset} & \textsc{E2E} \\
\midrule
\multicolumn{2}{c}{\textbf{\textsc{Natural Language Generation Details}}} \\
\midrule
\textsc{Top-p} & \textsc{0.9} \\
\textsc{Temperature} & \textsc{1e-12} \\
\textsc{Max new tokens} & \textsc{100} \\
\bottomrule
\end{tabular}
\end{center}
\end{table}

To demonstrate that fine-tuning is still necessary in specific domains and to show the effectiveness of our method in finding the best model under these circumstances, we conducted evaluations using the Korean Medical Multiple Choice Question Answering (KorMedMCQA) dataset~\citep{kweon2024kormedmcqa}. For batch learning, we used text segments with a maximum sequence length not exceeding 512 tokens. Consequently, the train, test, and validation sets for doctors contained 1,890, 285, and 164 examples, respectively; for nurses, the train, test, and validation sets contained 582, 291, and 291 examples; and for pharmacists, the train, test, and validation sets contained 692, 614, and 300 examples, respectively. For in-context learning, we provided examples within this length limit, and for classification fine-tuning, we used a linear head. For this, we used the \glspl{llama38b} model, the latest multilingual open-source large language model. This version was downloaded from this link.\footnote{\url{https://huggingface.co/meta-llama/Meta-Llama-3-8B-Instruct}}
More specific details about the model and experiments can be found in~\cref{llama3-exp-details}.

\begin{table}[ht]
\caption{Detailed \gls{llama38b} experimental setup.}
\label{llama3-exp-details}
\renewcommand{\arraystretch}{1.5}
\begin{center}
\setlength{\tabcolsep}{2pt}
\begin{tabular}{>{\raggedright\arraybackslash}p{0.3\linewidth} p{0.65\linewidth}}
\toprule
\textbf\textsc{{Category}} & \textbf{\textsc{Details}} \\
\midrule
\multicolumn{2}{c}{\textbf{\textsc{Model Specifications}}} \\
\midrule
\textsc{Architecture} & \textsc{Transformer} \\
\textsc{Pre-training} & \textsc{LLaMA-3-8B-Instruction} \\
\textsc{LoRA alpha} & \textsc{16} \\
\textsc{LoRA dropout} & \textsc{0.0} \\
\textsc{Optimizer} & \textsc{AdamW} \\
\textsc{Learning rate} & \textsc{[1e-06, 1e-03]} \\
\textsc{Batch size} & \textsc{[8, 16]} \\
\textsc{Gradient accumulation step} & \textsc{2} \\
\textsc{Epochs} & \textsc{10} \\
\midrule
\multicolumn{2}{c}{\textbf{\textsc{Task Specifications}}} \\
\midrule
\textbf{\textsc{Task name}} & \textsc{Korean Medical Question Answering} \\
\textsc{Prompt} & \begin{minipage}[t]{0.95\linewidth}
다음은 의사 면허 시험의 의료 질문입니다.
\\
질문을 읽고 올바른 답을 선택하세요.
\\
\\
항문압 측정 검사에서 항문 압력이 증가하는 경우는?

\begin{enumerate}[label=\Alph*.]
    \item 직장질루 \textsc{(rectovaginal fistula)}
    \item 항문열창 \textsc{(anal fissure)}
    \item 대변실금 \textsc{(fecal incontinence)}
    \item 대변메막힘 \textsc{(fecal impaction)}
    \item 직장탈출증 \textsc{(rectal prolapse)}
\end{enumerate}
답:
\\
\end{minipage} \\
\textsc{Dataset} & \textsc{KorMedMCQA} \\
\bottomrule
\end{tabular}
\end{center}
\end{table}

\subsection{Bayesian Optimization}

\paragraph{Details of HPBO.}
In the HPBO experiments, the number of iterations varied depending on the size of each dataset. Specifically, 20 iterations were conducted for the RTE dataset, while 10 iterations were carried out for the MRPC, SST2, and QNLI datasets. For the QQP and MNLI datasets, 8 iterations were performed. In addition, the SQuAD 2.0, SAMSum, and E2E datasets each underwent 10 iterations. These iteration counts were determined based on the respective sizes of the datasets. For single metric tasks, the chosen objective was the single valid metric itself. Conversely, for multi-metric tasks, the objective was the sum of all valid metrics.

\paragraph{Details of Sampling Fusion Members.}
We collected fusion members at step intervals ranging from 0.5 to 2.0 times the point of convergence identified in the training trajectory which represented $B$ in \cref{algorithm/ours}, adjusting the process to yield approximately 15 members in total. Additionally, for the \gls{roberta} model, we employed PyTorch's official \gls{swa} scheduler with cosine annealing. For the T5 and \gls{llama} models, we do not use any additional scheduler for collecting \gls{swa} members.

\paragraph{Details of MOBO.} 
In the case of \gls{mobo}, we initially provided the length of the fusion member and conducted iterations five times the total number of fusion members. This approach follows the common practice in BO of determining the initial points and the number of iterations based on the input dimension, allowing for the option to perform more iterations for improved performance. Furthermore, due to the differing scales of the loss and each metric, we applied min-max normalization to adjust the scales, utilizing the lowest value of single model performance obtained after the convergence point in the trajectory collection of members, rounded to the nearest value, as the minimum. The maximum values were determined by adding 0.1 for metric and 1.0 for loss respectively to this minimum value for use as the maximum. If there is a more critical metric or criterion, one can freely modify the optimization by placing weights according to the user's intent. No additional weights were applied in our experiments.
\section{Proposed Algorithm}
\label{app:sec:algorithm}

\begin{algorithm}[t]
\caption{Bayesian Optimization Model Fusion}
\label{algorithm/ours}
\begin{algorithmic}[1]
    \REQUIRE{Training set $\calD_{\text{trn}}$, validation set $\calD_{\text{val}}$, initial pre-trained weights $\btheta_{\text{init}}$, initial hyperparameters $\blambda_{\text{init}}$.}
    \ENSURE{Optimized hyperparameters $\blambda^*$, combination coefficients $\bdelta^*$.}
    \item[]
    \STATE{Initialize model $\calM$ with $\theta_{\text{init}}$, {\color{blue} Optionally freeze layers and cache intermediate features.}}
    \STATE{Initialize \gls{bo} with \gls{gp} model, starting with $\blambda_{\text{init}}$ \\
    and prior data $\calH_0 = (\blambda_{\text{init}}, \sum_{k=1}^{K} \Bar{f}_{\text{metric}}^{(k)}(\calM(\btheta_{\text{init}},\bw(\blambda_{\text{init}})), \calD_{\text{val}}))$.}
    \FOR{$i = 1$ {\bfseries to} $I$ iter}
        \STATE{Define LogEI using current \gls{gp}.}
        \STATE{Find $\blambda_i$ by optimizing LogEI.} 
        \STATE{Training $\calM (\btheta_{\text{init}}, \bw)$ with ($\calD_{\text{trn}}$, $\blambda_i$).}
        \STATE{Evaluate $\sum_{k=1}^{K} \Bar{f}_{\text{metric}}^{(k)}(\calM(\btheta_{\text{init}},\bw(\blambda_{i})), \calD_{\text{val}})$.}
        \STATE{Update \gls{gp} model with new data $(\blambda_i,\sum_{k=1}^{K} \Bar{f}_{\text{metric}}^{(k)}(\calM(\btheta_{i},\bw(\blambda_{i})), \calD_{\text{val}}))$.}
    \ENDFOR
    \STATE{Collect $\blambda^* = \argmin_{\lambda} \sum_{k=1}^K \Bar{f}_{\text{metric}}^{(k)}(\calM(\btheta_{\text{init}}, \bw(\blambda)), \calD_{\text{val}})$} 
    \STATE{$B^* \leftarrow \argmin_{B} \Bar{f}_{\text{metric}}^{(k)}(\calM(\btheta_{\text{init}},\bw_{B}(\blambda^*)), \calD_{\text{val}}) $}
    \STATE{$\calS \leftarrow \{\}$.}
    \FOR{$j = 1$ {\bfseries to} $J$ step}
        \STATE{Optimize $\calM (\btheta_{\text{init}}, \bw_j)$ with ($\calD_{\text{trn}}$, $\blambda^*$).}
        \IF {$j \geq 0.5B^*$}
        \STATE $\calS \leftarrow \calS \cup \calM (\btheta_{\text{init}}, \bw_j)$ 
        \ENDIF
    \ENDFOR
    \STATE Set reference point $\br$.
    \STATE $\calL_{\text{init}} \leftarrow$ $\{\Bar{f}_{\text{loss}}(\calM(\btheta_{\text{init}}, \Bar{\bw}(\bdelta_{\text{init}})), \calD_{\text{val}}), \Bar{f}_{\text{metric}}^{(1)}(\calM(\btheta_{\text{init}},\Bar{\bw}(\bdelta_{\text{init}})), \calD_{\text{val}}), \cdots$ \\
    $,\Bar{f}_{\text{metric}}^{(K)}(\calM(\btheta_{\text{init}},\Bar{\bw}(\bdelta_{\text{init}})), \calD_{\text{val}})\}$
    
    \STATE Initialize \gls{mobo} with \gls{gp} models, starting with $\Bar{\bw}_{\text{init}}$ and prior data $\mathcal{H}_0 = (\Bar{\bw}_{\text{init}}, \bdelta_{\text{init}}, \calL_{\text{init}})$.
    \STATE Compute initial Pareto optimal set $\mathcal{P}_0$ using $\mathcal{H}_0$.
    \FOR{$m = 1$ {\bfseries to} $M$ iter}
        \STATE{Define EHVI using current \gls{gp}s.}
        \STATE{Find $\bdelta_{m}$ by optimizing EHVI. \cref{ehvi}}
        \STATE{$\calL\leftarrow$ $\{\Bar{f}_{\text{loss}}(\calM(\btheta_{\text{init}},\Bar{\bw}(\bdelta_{m})), \calD_{\text{val}})
        , \Bar{f}_{\text{metric}}^{(1)}
        (\calM(\btheta_{\text{init}},\Bar{\bw}(\bdelta_{m})), \calD_{\text{val}}), \cdots$\\
        $, \Bar{f}_{\text{metric}}^{(K)}(\calM(\btheta_{\text{init}},\Bar{\bw}(\bdelta_{m})), \calD_{\text{val}})\}$}
        \STATE{Update \gls{gp}s with new data $(\bdelta_m,$$\calL)$} 
        \STATE{Update $\mathcal{P}_m$.}
    \ENDFOR
    \STATE{Collect $\bdelta^* =  \argmax_{\bdelta}\, \text{EHVI}(\bdelta; \calL, \br) $} 
\end{algorithmic}
\end{algorithm}

In this section, we provide an overview of the complete process of \gls{ours} encapsulated in the algorithm. \cref{algorithm/ours} systematically incorporates all three steps: 1) hyperparameter search via \gls{bo} as detailed in \cref{main:subsec:hyperparam}, 2) fusion member sampling as outlined in \cref{main:subsec:fusion_members}, and 3) identification of optimal $\bdelta^\star$ and model fusion through \gls{mobo} in \cref{main:subsec:MOBO}.
\begin{figure}[t]
    \centering
    \begin{subfigure}[b]{0.48\textwidth}
        \includegraphics[width=\textwidth]{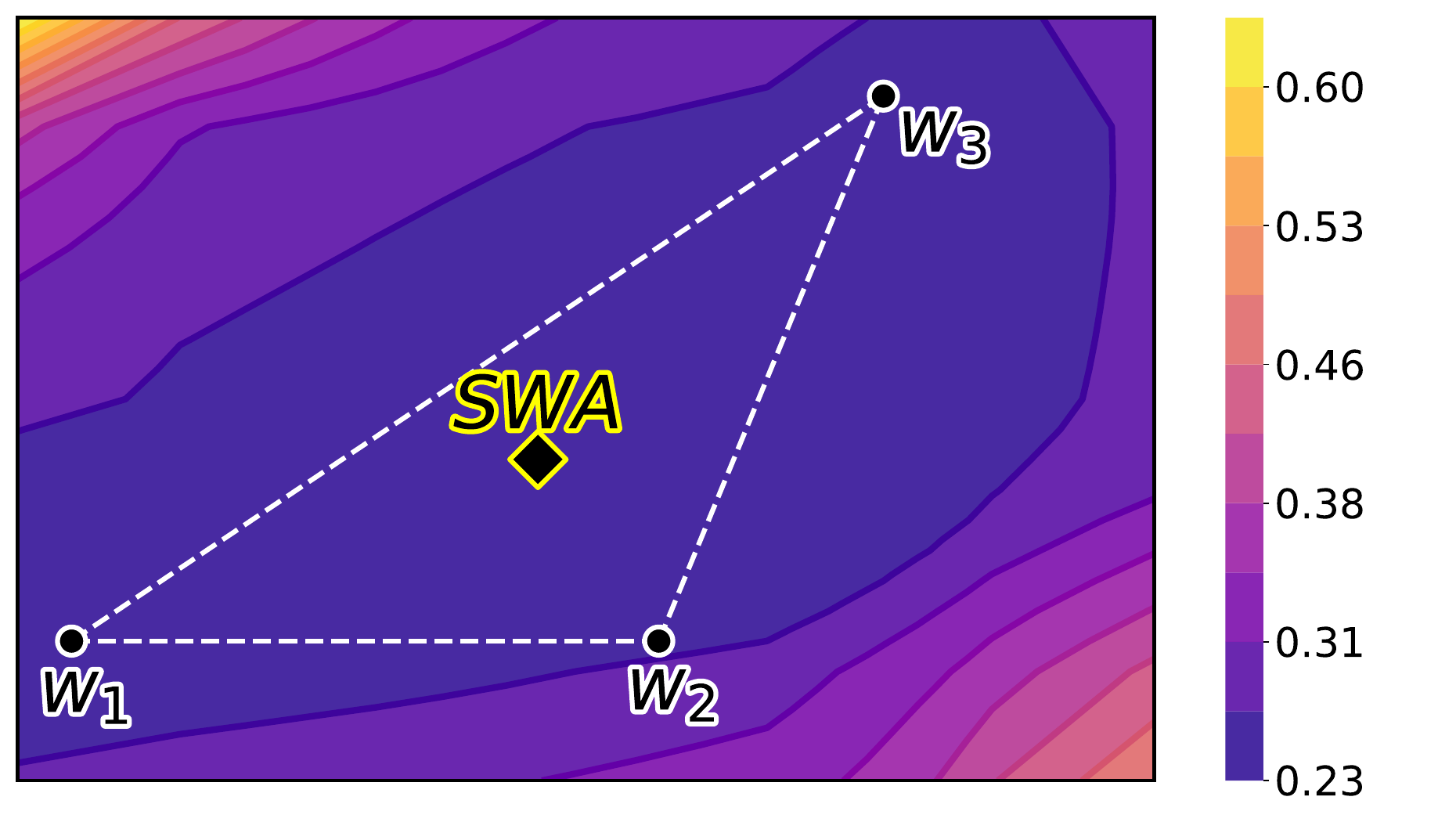}
        \caption{Loss surface of ViT-B/16.}
        \label{main:fig:vit_loss}
    \end{subfigure}
    \hfill
    \begin{subfigure}[b]{0.48\textwidth}
        \includegraphics[width=\textwidth]{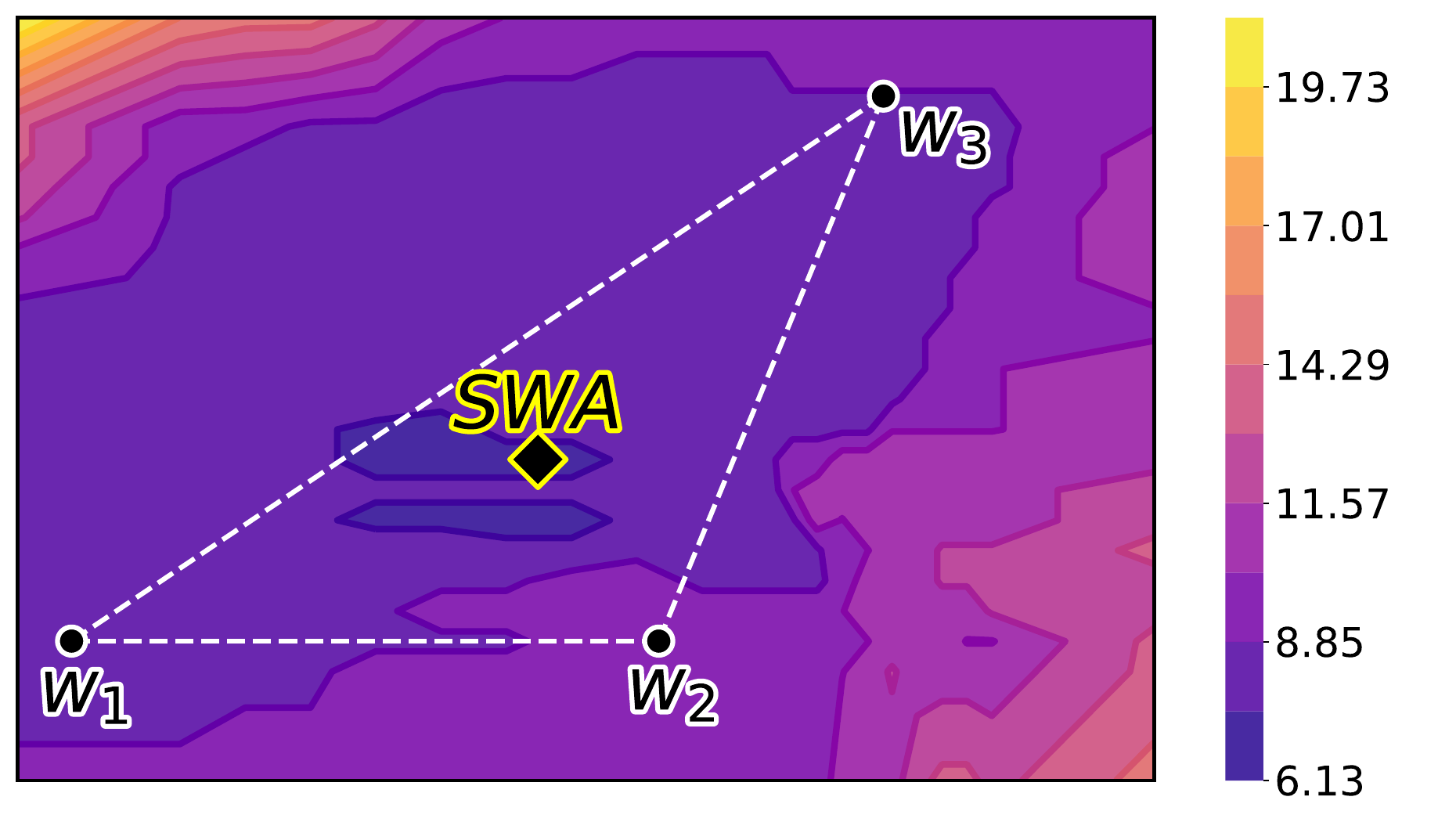}
        \caption{Metric surface of ViT-B/16.}
        \label{main:fig:vit_metric}
    \end{subfigure}

    \caption{Visualization of the loss landscape over parameters (a) and the metric landscape over parameters (b) for the vision task. The metric is accuracy error. We fine-tune the ImageNet-21k pre-trained ViT-B/16~\citep{dosovitskiy2021an} model on the Caltech-101 dataset. The members of the \gls{swa} for each figure are denoted as $w_1, w_2, w_3$. Here we can see similar trends with the ResNet-50 case.
    }
    \label{fig:vit_loss_metric_surfaces}
    \vspace{-3mm}
\end{figure}

\begin{table*}[t]
\centering
\caption{Spearman’s rank correlation coefficient value of (a) ResNet18 model on the CIFAR10 dataset, (b) ViT-B/16 model on the Caltech101 dataset, (c) \gls{roberta}-base model on the SST-2 dataset, (d) T5-base model on the SQuAD2.0 dataset, and (e) \gls{llama27b} model on the E2E dataset. Here we used 15 fine-tuned weights for each task to measure the Spearman’s rank correlation. A higher value indicates a higher correlation between metric and loss value.}
\label{tab:spearman-correlation}
\begin{small}
\begin{sc}
\begin{tabular}{lccccc}
\toprule
Metric & (a) & (b) & (c) & (d) & (e) \\
\midrule
Spearman’s RCC & 0.6182 & 0.6558 & 0.1430 & 0.2150 & 0.2286 \\
\bottomrule
\end{tabular}
\end{sc}
\end{small}
\end{table*}

\section{Additional Experiments}
\label{app:sec:additional_experiment}

In this section we demonstrate additional experiments not included in the main article.

\subsection{Additional Experiments on Loss and Metric Landscapes}
\label{app:sec:additional_experiment_surface}

We also explored the potential loss metric discrepancy in \glspl{plm}, which might originate from the inherent features of transformer attention or from the use of adaptive optimizers. To analyze deeper, we visualize the loss-metric surface of the ViT-B/16 model per-trained on ImageNet-21k, using Adam optimizer which is the same as the optimizer of our language model.
According to \cref{fig:vit_loss_metric_surfaces}, unlike in \glspl{plm},
the optimal points for loss and performance metrics in the ViT-B/16 model were aligned, indicating a distinct behavior between language and vision transformers in this context. Additionally, we have also undertaken the measurement of Spearman's rank correlation between loss and metric values across a range of models and tasks in both \gls{nlp} and CV domains.
\cref{tab:spearman-correlation} clearly shows that the correlations in \gls{nlp} tasks are less than in CV tasks.

\begin{figure}[ht]
    \centering
    \begin{subfigure}[b]{0.48\textwidth}
        \includegraphics[width=\textwidth]{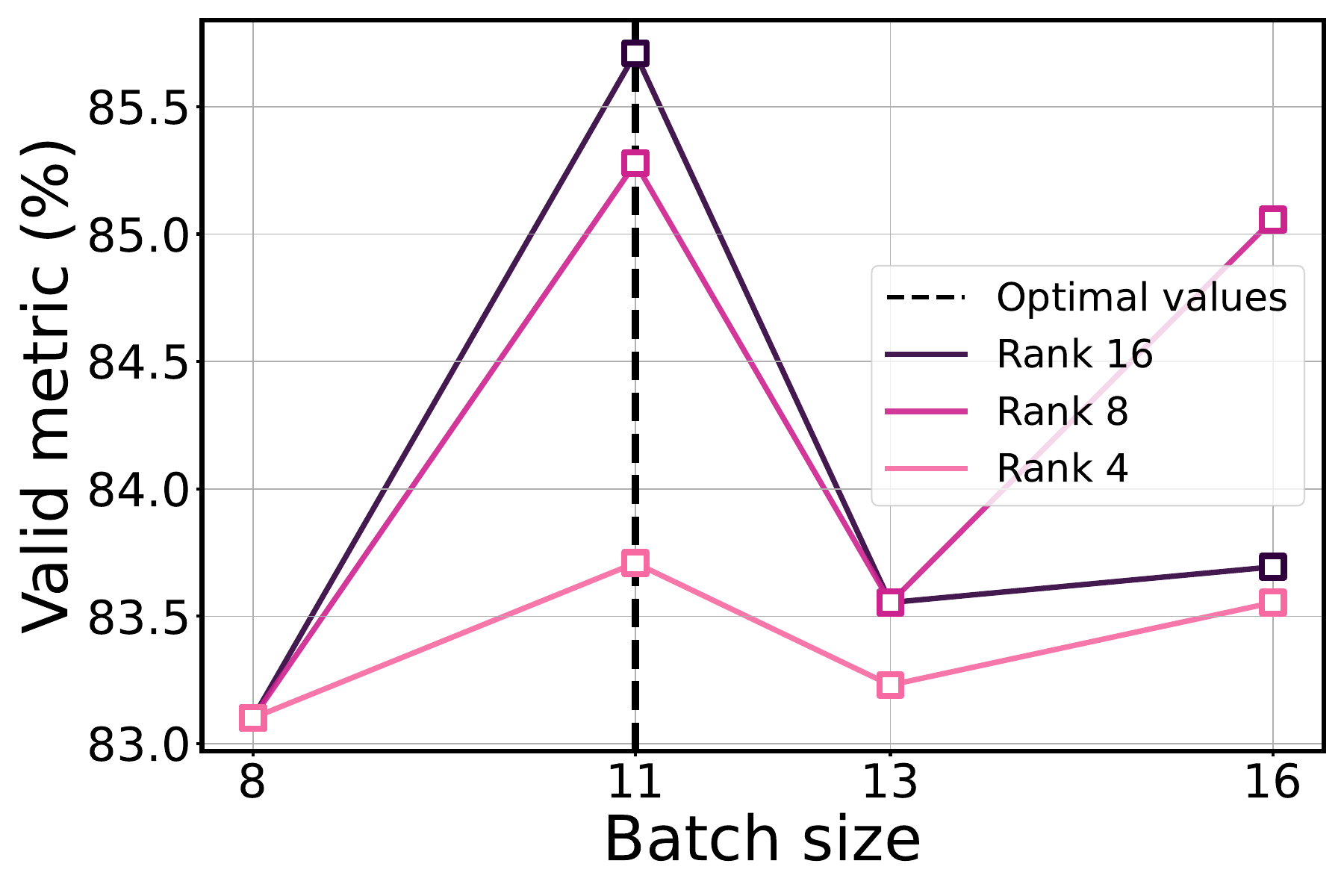}
        \caption{Validation metric results for the varying batch size and the number of LoRA rank.}
        \label{main:fig:lora_loss}
    \end{subfigure}
    \hfill
    \begin{subfigure}[b]{0.48\textwidth}
        \includegraphics[width=\textwidth]{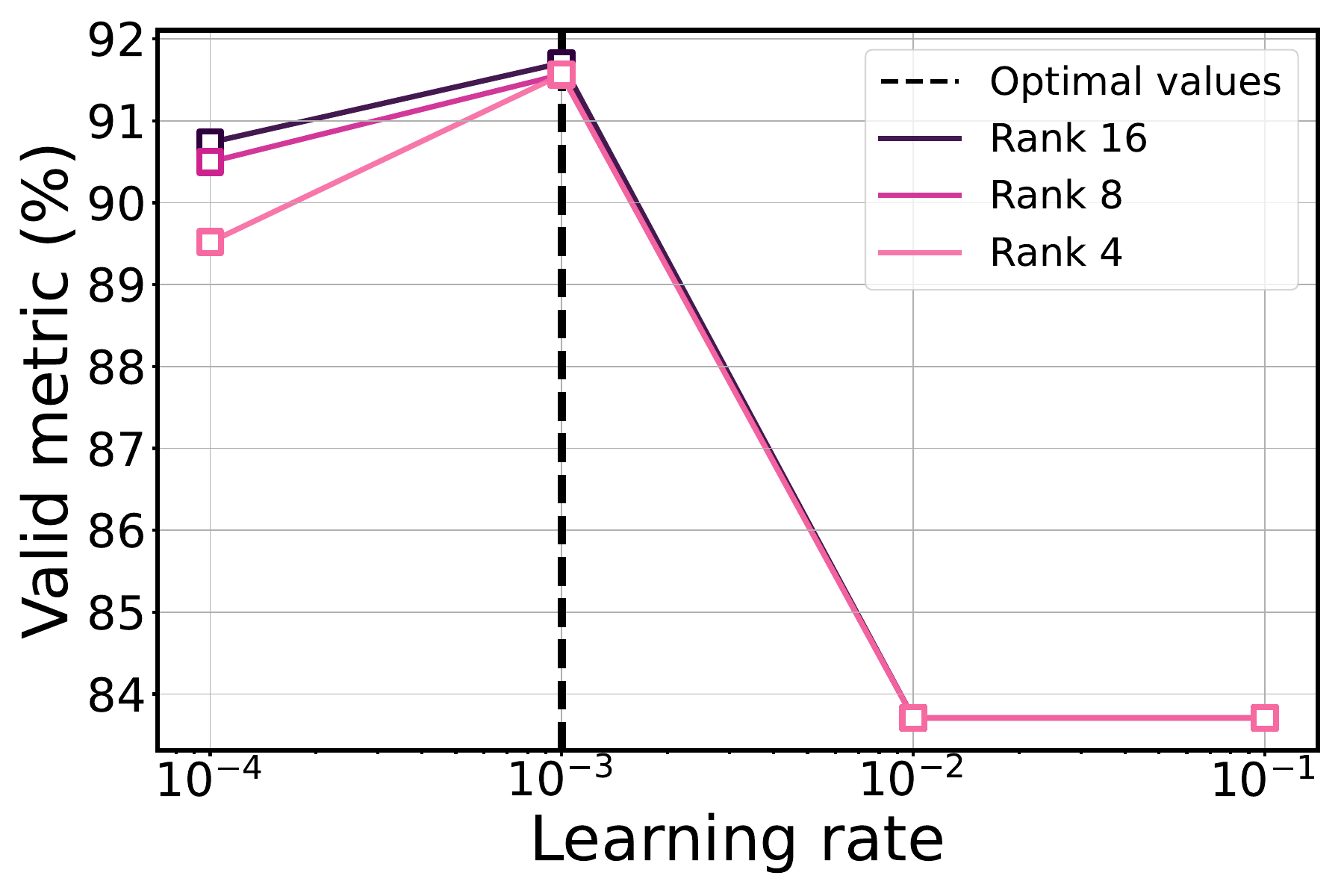}
        \caption{Validatoin metric results for the varying learning rate and the number of LoRA rank.}
        \label{main:fig:lora_metric}
    \end{subfigure}

    \caption{Validation loss and metric (F1 score) results for the varying hyperparameter ((a) batch size, (b) learning rate) and the number of \gls{lora} rank for the \gls{roberta} on MRPC dataset. (a) and (b) indicate that the optimal hyperparameters consistently align well across different numbers of \gls{lora} rank.}
    \label{app:fig:lora_alignment}
\end{figure}

\begin{figure}[ht]
    \centering
    \begin{subfigure}[b]{0.48\textwidth}
        \includegraphics[width=\textwidth]{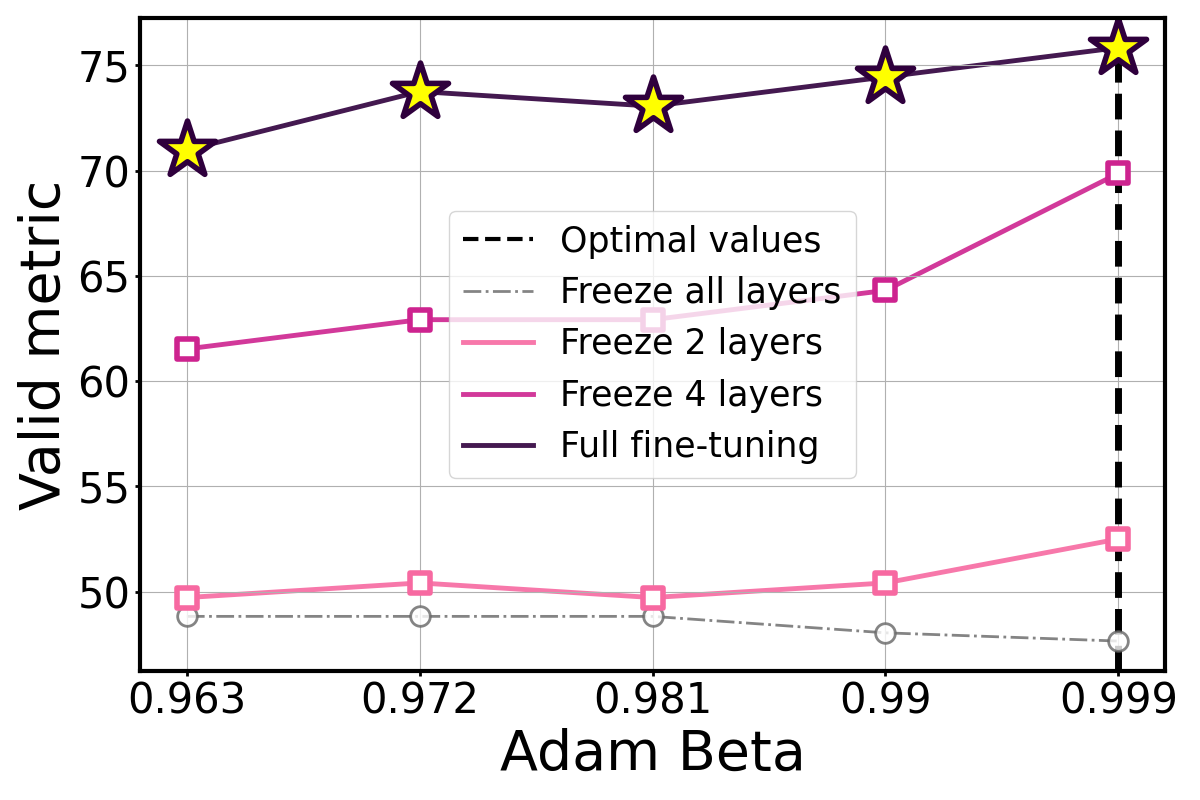}
        \caption{Validation metric (Accuracy) for varying beta parameters and frozen layers.}
        \label{fig:R3}
    \end{subfigure}
    \hfill
    \begin{subfigure}[b]{0.48\textwidth}
        \includegraphics[width=\textwidth]{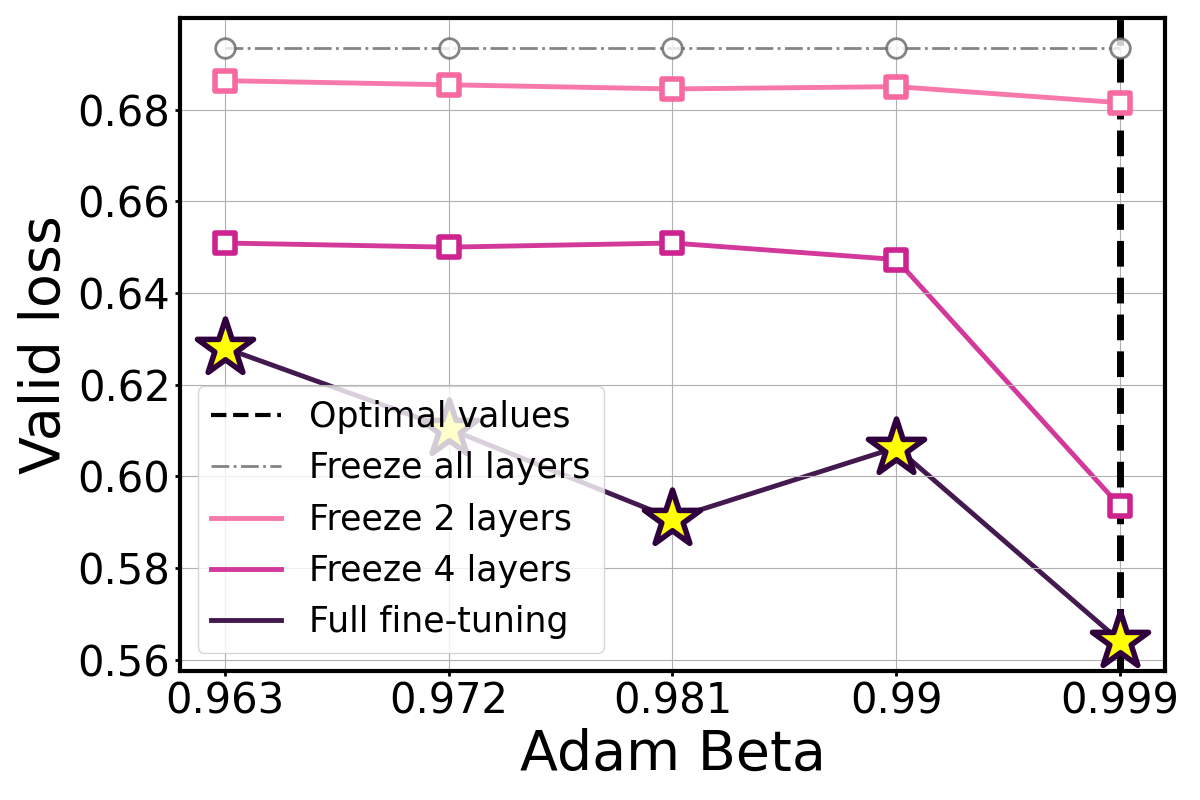}
        \caption{Validation loss for varying beta parameters and frozen layers.}
        \label{fig:R4}
    \end{subfigure}
    \caption{Results for the \gls{roberta}-base model on the RTE dataset. (a) and (b) indicate that the optimal hyperparameters align well across different numbers of frozen layers, except when all pre-trained layers are frozen.}
    \label{fig:R3_R4}
\end{figure}

\begin{figure}[ht]
    \centering
    \begin{subfigure}[b]{0.48\textwidth}
        \includegraphics[width=\textwidth]{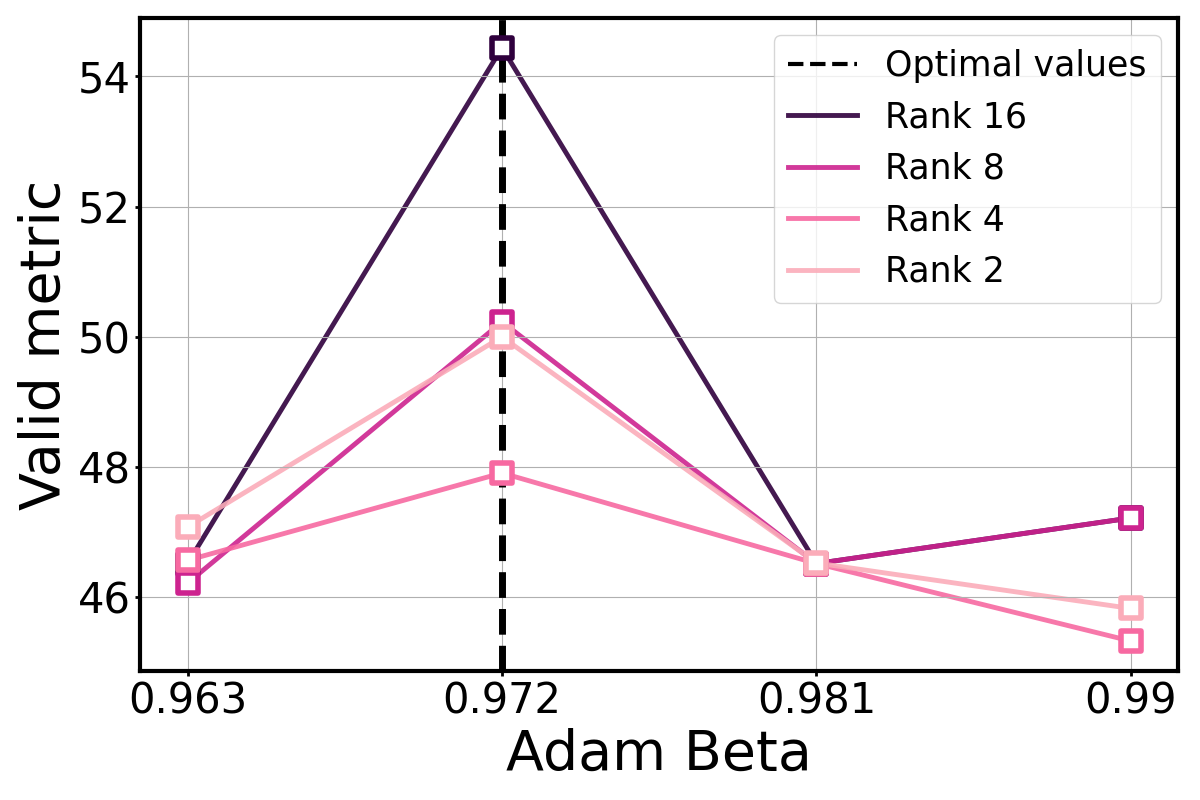}
        \caption{Validation metric (Accuracy) for varying beta parameters and LoRA ranks.}
        \label{fig:R5}
    \end{subfigure}
    \hfill
    \begin{subfigure}[b]{0.48\textwidth}
        \includegraphics[width=\textwidth]{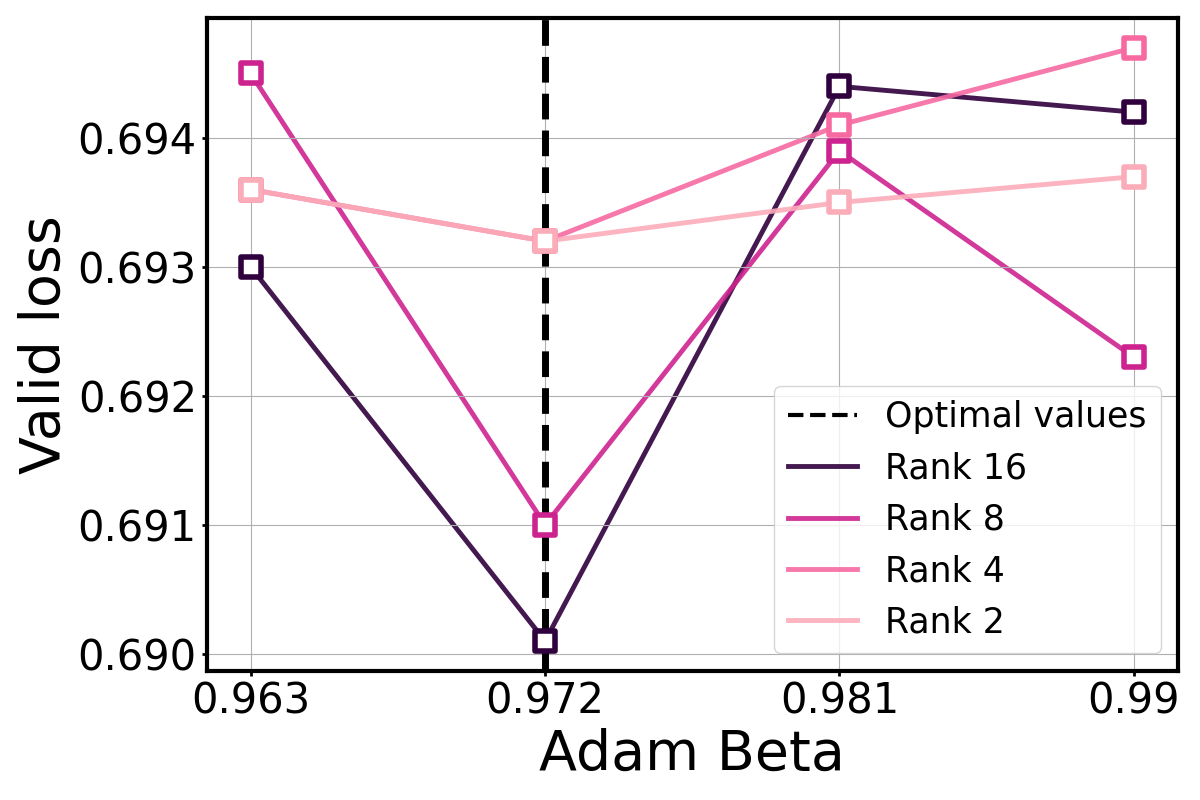}
        \caption{Validation loss for varying beta parameters and LoRA ranks.}
        \label{fig:R6}
    \end{subfigure}
    \caption{Results for the \gls{roberta}-base model on the RTE dataset. (a) and (b) highlight that optimal hyperparameters align well across different numbers of LoRA ranks, emphasizing the importance of parameter tuning.}
    \label{fig:R5_R6}
\end{figure}

\begin{figure}[ht]
    \centering
    \begin{subfigure}[b]{0.48\textwidth}
        \includegraphics[width=\textwidth]{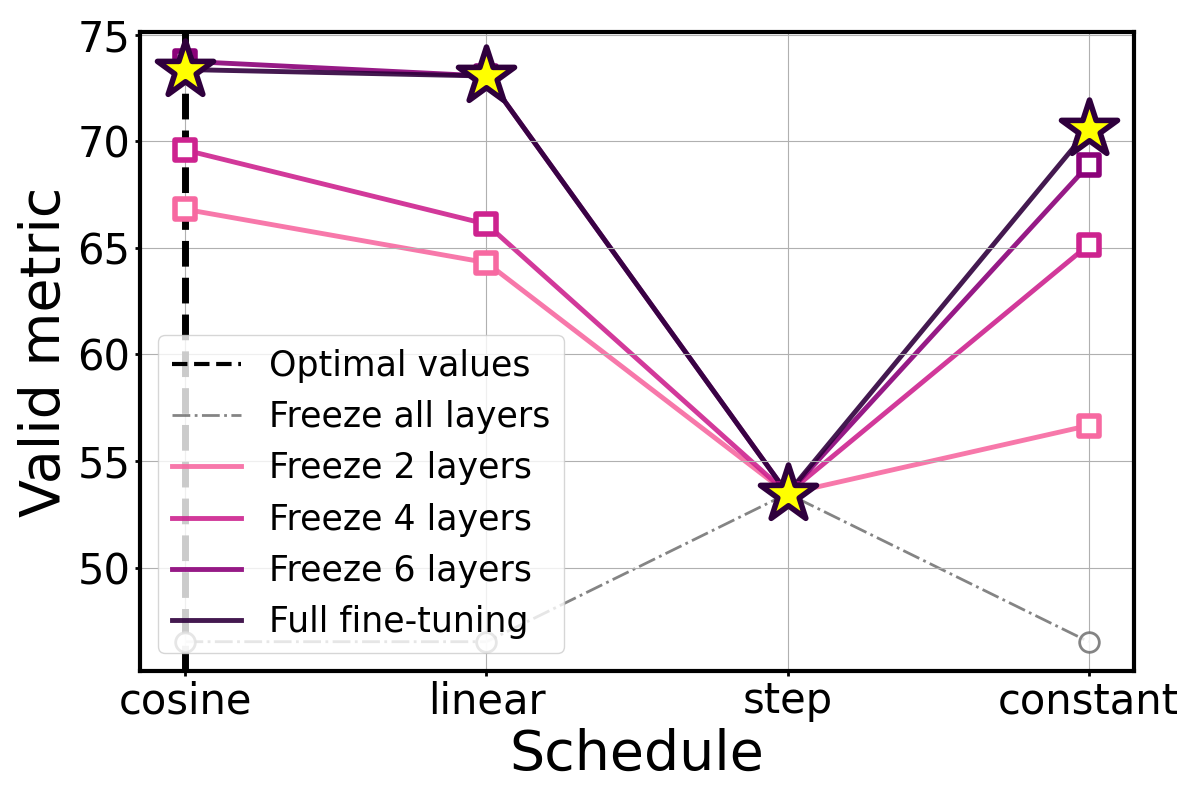}
        \caption{Validation metric (Accuracy) for varying learning rate schedule methods and frozen layers.}
        \label{fig:R7}
    \end{subfigure}
    \hfill
    \begin{subfigure}[b]{0.48\textwidth}
        \includegraphics[width=\textwidth]{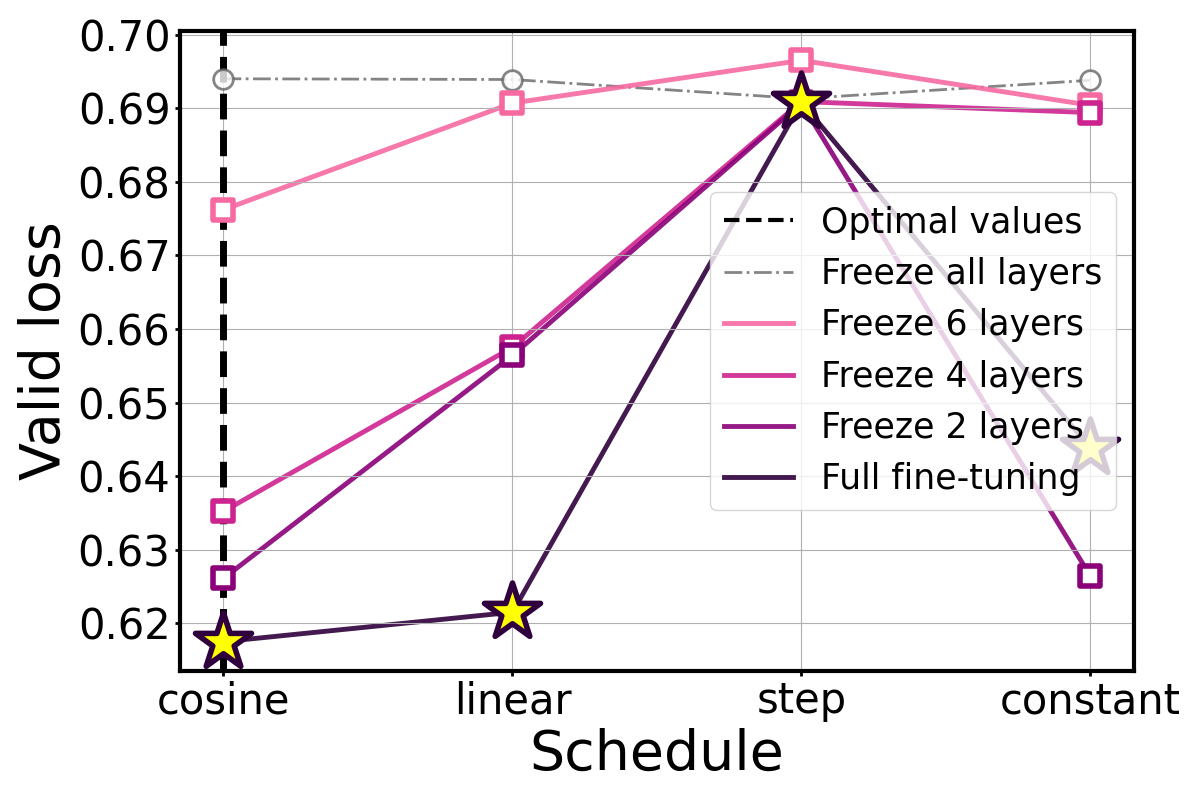}
        \caption{Validation loss for varying learning rate schedule methods and frozen layers.}
        \label{fig:R8}
    \end{subfigure}
    \caption{Results for the \gls{roberta}-base model on the RTE dataset. (a) and (b) demonstrate that optimal hyperparameters are consistent across different numbers of frozen layers, indicating the critical role of hyperparameter choices.}
    \label{fig:R7_R8}
\end{figure}

\subsection{Additional Experiments on Hyperparameter Alignment}
\label{app:hp_align}

In our experiments utilizing \gls{lora}, we aimed to verify whether the optimal hyperparameters align when using a smaller rank to reduce costs, compared to using a larger rank. \cref{app:fig:lora_alignment} demonstrates that the optimal batch size and learning rate align even when the number of LoRA ranks varies. In addition to these hyperparameters, we investigated the potential alignment of the beta parameter of the Adam optimizer, which is the standard optimizer for training large language models, as well as the scheduler.

\cref{fig:R3_R4} and \cref{fig:R5_R6} indicate that, with the exception of the scenario where all layers are frozen, the optimal beta parameter of the Adam optimizer consistently aligns regardless of the number of frozen layers or the LoRA rank. \cref{fig:R1} and \cref{fig:R2} suggest that the optimal points of different learning rate schedules can be aligned according to the number of frozen layers.

These results indicate that by employing a lightweight model, we can identify the optimal hyperparameters, thereby simplifying the hyperparameter optimization process and reducing computational time.

\subsection{Additional Ablation Experiments}

\paragraph{Metric Function in Multi-Objective Bayesian Optimization.}

\begin{figure}[ht]
    \centering
    \includegraphics[width=0.6\textwidth]{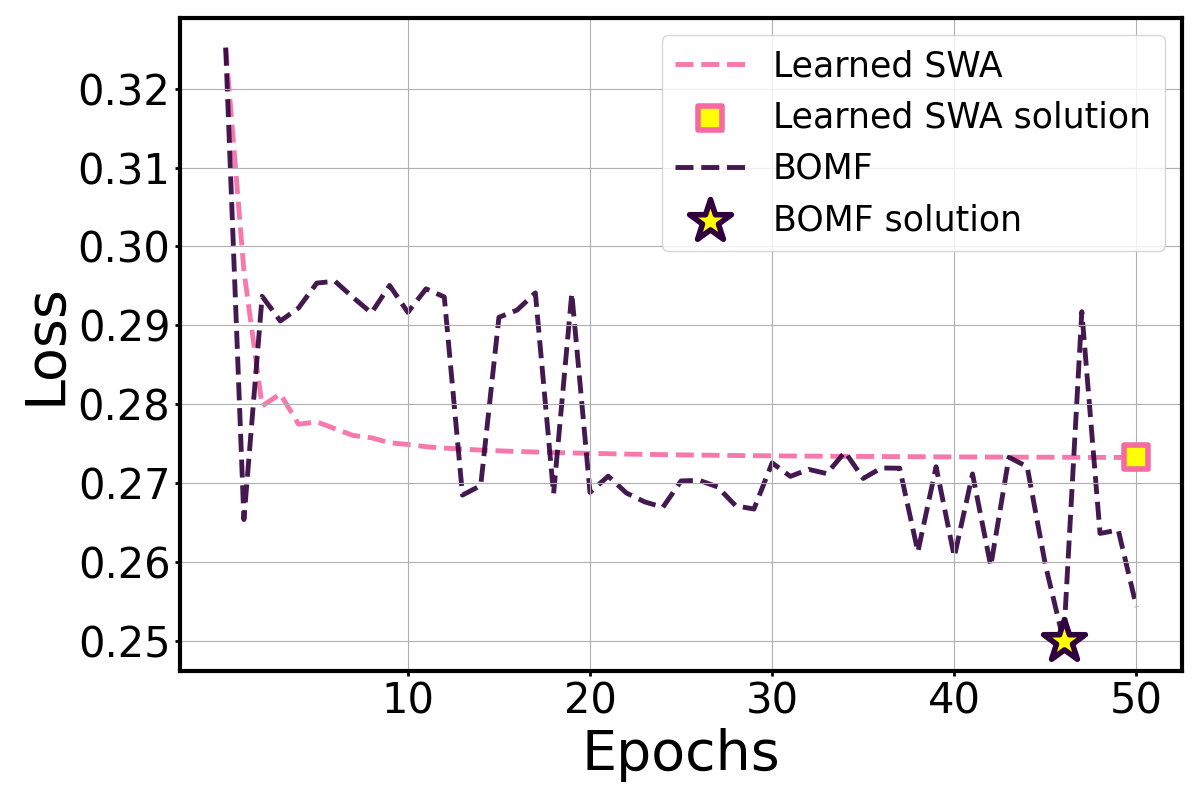}
    \caption{Visualization of the evolution of loss throughout the optimization phases. We conduct experiments on the RTE dataset with the \gls{roberta}-base model. Learned \gls{swa}, an optimization process without metrics, tends to converge to local minima close to the starting point. Conversely, \gls{ours} exhibits a more successful exploration, incorporating metrics, and ultimately discovers a more robust solution.}
    \label{fig:R1}
\end{figure}

\begin{figure}[ht]
    \centering
    \includegraphics[width=0.6\textwidth]{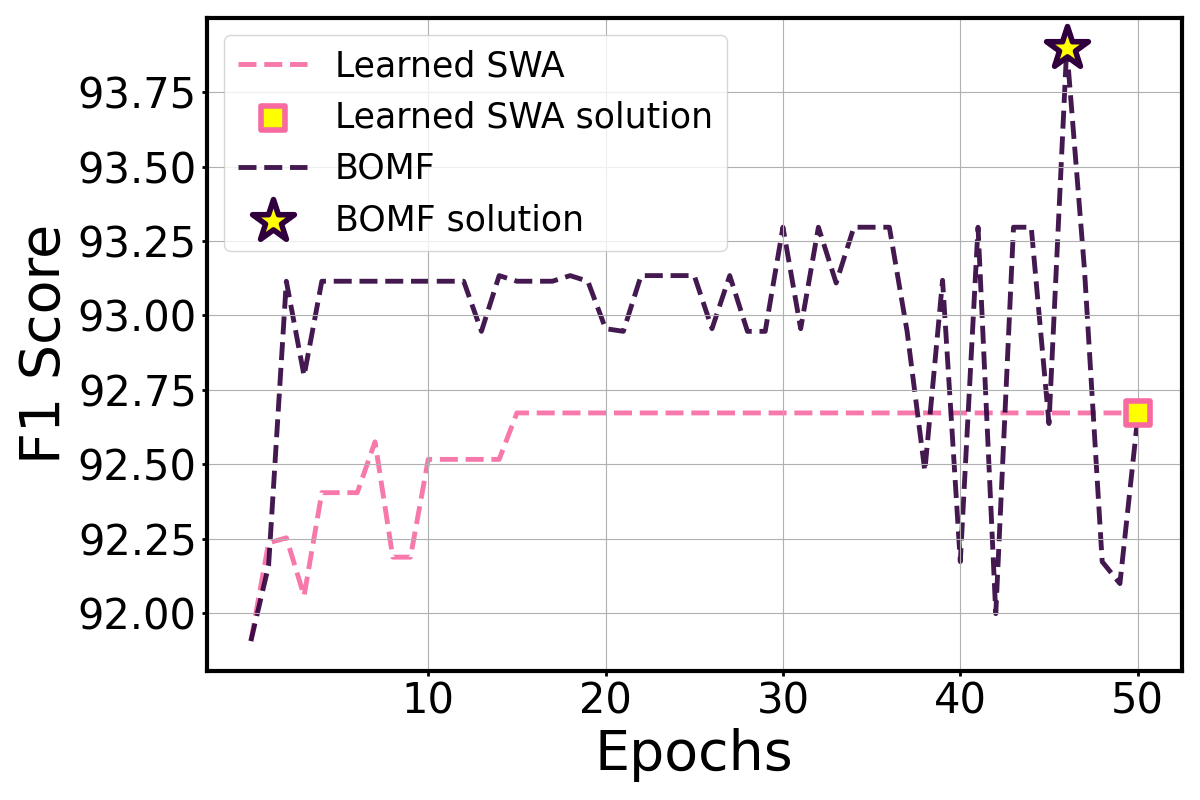}
    \caption{Visualization of the evolution of metric throughout the optimization phases. We conduct experiments on the RTE dataset with the \gls{roberta}-base model. Like the loss, Learned \gls{swa} tends to converge towards local minima without enhancing metric performance after the convergence. In contrast, \gls{ours} effectively navigates through explorations to discover a high-performing solution.}
    \label{fig:R2}
\end{figure}

In the \cref{main:sec:mismatch}, we pointed out the issues of existing methods that perform fusion based solely on loss, particularly due to the discrepancy between metric and loss in language models. We demonstrated that our method outperforms the learned-\gls{swa} method, which relies only on loss. To strengthen our claim, we additionally compared the loss and metric values between optimization processes with and without metrics. Specifically, we examined the loss and metric values of weights along the training trajectory between the initial point and the optimized point of Learned SWA and BOMF.
\cref{fig:R1,fig:R2} show that BOMF generates complex trajectories for both loss and metric, exploring solutions based on both criteria. In contrast, Learned \gls{swa}, relying solely on the loss function, gets trapped in local minima around the starting point, failing to discover the optimal solution. This suggests that BOMF's exploration property and the inclusion of metrics help escape local minima and discover more robust solutions.

\paragraph{Loss Function in Multi-Objective Bayesian Optimization.}
\label{app:ablation}

\begin{table*}[t]
\centering
\caption{Spearman’s rank correlation coefficient value between loss and metric for the various optimization processes in the middle scale tasks. Here, we assess the correlation of the \gls{roberta}-base model on the MRPC dataset (a) and the RTE dataset (b). Additionally, we examine the correlation between loss and metric of the T5-base model on the SQuAD2.0 dataset using the F1 score (c) and the EM (d) metrics. We conduct evaluations on a total of 100 sampled sets for (a) and (b), and 30 for (c) and (d).  Here, Loss \gls{bo} \gls{swa} and Metric \gls{bo} \gls{swa} denote the approach where we exclusively employ either the loss function or the metric during the \gls{mobo} process, respectively.
}
\label{tab:middle_scale_tasks_spc}
\begin{small}
\begin{sc}
\begin{tabular}{lcccc}
\toprule
Metric & (a) & (b) & (c) & (d) \\
\midrule
Baseline (HPBO) & 0.1640 & 0.3256 & 0.2076 & 0.3021 \\
Loss BO SWA & 0.1903 & 0.4334 & 0.4210 & 0.4710 \\
Metric BO SWA & 0.2465 & 0.3755 & 0.5333 & 0.5000 \\
BOMF & 0.5189 & 0.5886 & 0.6500 & 0.6000 \\
\bottomrule
\end{tabular}
\end{sc}
\end{small}
\end{table*}

\begin{table*}[t]
\centering
\caption{Spearman’s rank correlation coefficient value between loss and metric for the various optimization processes in large-scale tasks. We evaluate the \gls{llama27b} on the SAMSum dataset. Here, Loss \gls{bo} \gls{swa} and Metric \gls{bo} \gls{swa} denote the approach where we exclusively employ either the loss function or the metric during the \gls{mobo} process, respectively.
}
\label{tab:large_scale_tasks_spc}
\begin{small}
\begin{sc}
\begin{tabular}{lcccc}
\toprule
Metric & R1 & R2 & RL & Avg. \\
\midrule
Baseline (HPBO) & 0.4024 & 0.2000 & 0.1030 & 0.2351 \\
Loss BO SWA & 0.5335 & 0.1306 & 0.1197 & 0.2613 \\
Metric BO SWA & 0.5330 & 0.6060 & 0.6060 & 0.5817 \\
BOMF & 0.6445 & 0.6152 & 0.6137 & 0.6245 \\
\bottomrule
\end{tabular}
\end{sc}
\end{small}
\end{table*}

\gls{ours} adopts both loss and metric as objectives when optimizing $\bdelta$. This approach is based on the understanding that the loss provides a macroscopic guide for overall metric performance. As seen in \cref{fig:loss_metric_surfaces}, the solution of \gls{swa} is optimal on the loss surface but not on the metric surface. In contrast, \gls{ours} performs well on both the loss and metric surfaces. These results indicate that \gls{ours} ensures a high correlation between loss and metric values, preventing overfitting to either loss or metric during validation and enhancing the robustness of the weights on the test dataset. This clearly shows that including the loss function provides more useful guidance than optimizing $\bdelta$ exclusively in a complex and sharp metric landscape.

Moreover,
as demonstrated in~\cref{tab:middle_scale_tasks_spc,tab:large_scale_tasks_spc},
our approach, which utilizes both loss and metric, improves the correlation between them. This leads to reaching the point of optimal performance,
as shown in \cref{tab:mid_exp1_full,tab:mid_exp2_full,tab:large-exp2_full}.
These findings support that \gls{ours} outperforms Bayesian optimization using a single metric without the loss function.

\paragraph{Multiple Metric Functions in Multi-Objective Bayesian Optimization.}
For tasks with multiple metrics, we optimized using all the available metrics. Therefore, it is important to investigate how multiple metrics impact the optimization process. In \cref{tab:ab:single-multi}, we validated this through performance evaluations, but we also assessed changes in correlation. \cref{tab:large_scale_tasks_spc} shows that optimizing with a single metric can weaken the correlation between the loss and other metrics.

\paragraph{Impact of the Better Trajectories in Model Fusion Performance}
\gls{ours} aims to construct the best-performing single model through model fusion within a parameter space. As detailed in \cref{main:subsec:fusion_members}, different combinations of fine-tuning hyperparameters (such as learning rate and batch size) yield varied generalization performances after the fine-tuning process. Therefore, to identify the best-performing single model after the model fusion, we must first determine the optimal hyperparameters that yield the best-performing single model before the fusion. In \cref{tab:mid_exp1_full}, we can confirm that the performance of the final \gls{ours}, executed after finding good hyperparameters through outerbo, is better than the fusion performance executed with hyperparameters obtained from grid search.

\paragraph{Performance of Conventional Optimization Strategies for Each Task.}

\begin{table*}[ht!]
    \centering
    \caption{\textbf{Results of BOMF and Other Neural Network Optimization Strategies on Various Datasets.} A higher value is better for all the metrics.}
    \label{tab:nn_optimizer}
    \setlength{\tabcolsep}{3pt}
    \begin{small}
    \begin{sc}
    \begin{tabular}{lcccccc}
        \toprule
        & \multicolumn{3}{c}{GLUE (RoBERTa-base)} & \multicolumn{3}{c}{SAMSum (LLaMA2-7B)} \\
        \cmidrule(lr){2-4} \cmidrule(lr){5-7}
        Optimization Strategy & RTE (ACC) & MRPC (F1) & SST2 (ACC) & R1 & R2 & RL \\
        \midrule
        SGD & 54.37 & 88.90 & 90.11 & 50.63 & 19.34 & 41.24 \\
        AdamW & 78.49 & 92.72 & 94.41 & 52.25 & 27.52 & 44.03 \\
        SWA & 76.70 & 91.73 & 94.75 & 52.21 & 27.58 & 44.04 \\
        SAM & 77.34 & 92.56 & 94.77 & 52.75 & 28.55 & 44.16 \\
        BOMF & \textbf{81.40} & \textbf{93.90} & \textbf{95.54} & \textbf{53.07} & \textbf{28.61} & \textbf{44.40} \\
        \bottomrule
    \end{tabular}
    \end{sc}
    \end{small}
\end{table*}

Additionally, we conducted experiments to directly compare the original optimizers with \gls{ours}. We found the optimal hyperparameters for the original optimizers using \gls{bo} and then performed experiments on subsets of the GLUE dataset using the \gls{roberta} model and the SAMSum dataset using the \gls{llama27b} model. The results presented in \cref{tab:nn_optimizer} clearly demonstrate that \gls{ours} outperforms the best performance achievable with any of the original optimizers.

\paragraph{Additional Experiments with BOMF Using LLM-Based Evaluation}

To validate BOMF's performance under diverse evaluation metrics, we conducted experiments using a ChatGPT-3.5-Turbo-based approach. This method involves scoring by asking the LLM to assess the similarity between generated responses and ground-truth answers. Using this, BOMF was benchmarked against other models. As shown in \cref{tab:evaluation_chatgpt_3_5_turbo}, BOMF consistently outperformed these baselines, highlighting its capability to adapt not only to traditional metrics but also to newer evaluation techniques.

This robustness is further reinforced by BOMF's design, which combines loss with multiple metrics, enhancing its generalizability across unseen metrics. Notably, BOMF leverages BO for tuning combination coefficients, optimizing based on evaluation values rather than backward processing through the metrics themselves. This approach allows BOMF to efficiently optimize coefficients across complex evaluation settings, as shown in the ChatGPT BOMF column in \cref{tab:evaluation_chatgpt_3_5_turbo}.

These findings underscore BOMF's adaptability, demonstrating its resilience across varied evaluation frameworks, including those generated by LLMs.

\begin{table*}[ht]
\centering
\caption{\textbf{Evaluation Results Using ChatGPT-3.5-Turbo.} (a) Baseline, (b) SWA, (c) Greedy SWA, (d) Learned SWA, (e) BOMF, and (f) ChatGPT BOMF.}
\label{tab:evaluation_chatgpt_3_5_turbo}
\begin{sc}
    \begin{tabularx}{\columnwidth}{X}
    \toprule
    \textbf{Evaluation prompt example} \\
    \midrule
You are an automated grading assistant helping a teacher grade student answers.\\ 
\\
The correct summary for this text is: <ground-truth>\\
A student submitted the summary: <prediction>\\
\\
The student's summary must be correct and specific but not overcomplete. Small differences in formatting should not be penalized.
On a scale from 0 to 100, where 0 means completely incorrect and 100 means completely correct, how similar is the student's summary to the ground truth? Please provide only a numerical score without any explanation.\\
    \bottomrule
    \end{tabularx}
    \begin{tabularx}{\columnwidth}{r*{6}{X}}
        & \textbf{(a)} & \textbf{(b)} & \textbf{(c)} & \textbf{(d)} & \textbf{(e)} & \textbf{(f)} \\
        \midrule
        \textbf{Grade} & 70.74 & 71.92 & 72.05 & 71.80 & 72.64 & 73.18 \\
        \bottomrule
    \end{tabularx}
\end{sc}
\end{table*}

\begin{table*}[ht!]
\centering
\caption{\textbf{Full Results on Text Classification Task Using \gls{robertabase}.} Results of \gls{ours} and baseline methods with GLUE benchmark datasets. ACC and F1 denote metrics for each dataset, representing accuracy and F1 score, respectively.}
\label{tab:mid_exp1_full}
\begin{scriptsize}
\begin{sc}
\setlength{\tabcolsep}{4pt}
    \begin{tabular}{lccccccc}
\toprule
    & \multicolumn{6}{c}{Dataset} & \multirow{2}{*}{Avg.} \\
    \cmidrule(lr){2-7}
    Method & RTE ({\footnotesize Acc}) & MRPC ({\footnotesize F1}) & SST-2 ({\footnotesize Acc}) & QNLI ({\footnotesize Acc}) & QQP ({\footnotesize F1}) & MNLI ({\footnotesize Acc}) \\
\midrule
Grid Fine-tune & 77.78 & 92.39 & 94.87 & 92.62  & 88.16 & 87.41 & 88.93  \\
Grid Search Inner BO & 79.66 & 92.48 & 95.31 & 92.79 & 88.24 & 87.54 &\\
\midrule
HPBO (Full) & 78.57 & 92.78 & 95.11 & 93.01  & 88.58 & \UL{87.46} & 89.25  \\
SWA & 78.62 & 92.24 & 95.42 & 92.81  & 88.49 & 87.41 & 89.17  \\
Greedy SWA & 80.70 & 92.83 & \UL{95.54} & \UL{93.16}  & \UL{88.64} & 87.45 & \UL{89.72} \\
Learned SWA & \UL{81.40} & 92.81 & 95.31 & 92.94  & 88.38 & 87.41 & 89.71 \\
TWA & 81.23 & 91.58 & \UL{95.54} & 93.00 & 87.85 & 87.42 & 89.44  \\
BOMF (ours) & \BL{81.75} & \BL{93.37} & \BL{95.65} & \BL{94.83} & \BL{88.66} & \BL{87.51} & \BL{90.30}  \\
\midrule
HPBO (Freeze) & 78.49 & 92.72 & 94.41 & 92.71  & 88.04 & 87.45 & 88.97 \\
SWA & 76.70 & 91.73 & 94.75 & 93.21  & 88.35 & 87.44 & 88.70  \\
Greedy SWA & \UL{80.01} & 93.38 & \UL{95.20} & \UL{93.30}  & \UL{88.67} & \UL{87.84} & \UL{89.73}  \\
Learned SWA & 78.79 & \UL{93.56} & \UL{95.20} & 93.03  & 88.43 & 87.55 & 89.43 \\
TWA & 78.29 & 92.16 & 94.87 & 92.86 & 88.59 & 87.44 & 89.03  \\
BOMF (ours) & \BL{81.40} & \BL{93.90} & \BL{95.54} & \BL{93.50}  & \BL{88.68} & \BL{87.86} & \BL{90.15} \\
\bottomrule
\end{tabular}
\end{sc}
\end{scriptsize}
\end{table*}

\subsection{Full Experimental Results}
\label{app:full_exps}

In this section, we present full experimental results encompassing text classification tasks for the \gls{mlm} and question answering, summarization, and dialogue generation tasks for the autoregressive \gls{llm}. In all tables, the best performance is indicated with \BL{boldfaced underline}, and the second-best value is represented with \UL{underline} for methods that use the same best hyperparameters by \cref{main:subsec:hyperparam}. The final column `Avg.' provides a summary of overall results for each method across various datasets or metrics. The terms `Full' and `Freeze' in~\cref{tab:mid_exp1_full,tab:mid_exp2_full} specify the exploration of optimal hyperparameters using either the entire model or a model with half of its weights frozen, as discussed in~\cref{main:subsec:hyperparam}.
Similarly, the terms `Rank 64' and `Rank 4' in \cref{tab:large_exp1_full,tab:large-exp2_full} denote that we use the Rank 64 or the lightweight Rank 4 version of \gls{lora} model for the hyperparameter search, respectively.

\paragraph{Text Classification.}
\cref{tab:mid_exp1_full} demonstrates the consistently better performance of \gls{ours} over other baselines that employ the same best hyperparameters. These findings affirm the effectiveness of \gls{ours} in the context of single-metric \gls{nlp} tasks.

\paragraph{Question Answering.}
\cref{tab:mid_exp2_full} presents the complete experimental results for the question-answering task. \gls{ours} consistently surpasses other baselines utilizing the same best hyperparameters. These outcomes prove the effectiveness of \gls{ours} in the realm of multi-metric \gls{nlp} tasks, improving both F1 and EM metrics concurrently.

\paragraph{Summarization.}
\cref{tab:large_exp1_full} provides empirical evidence that \gls{ours} achieves the highest average performance across evaluated metrics. While Learned SWA and Greedy SWA exhibit the best performance results in R1 and R2 metrics, respectively, they experience declines across other metrics. However, our approach demonstrates a consistent improvement across all metrics. These results prove the efficacy of \gls{ours} in multi-metric \gls{nlp} tasks.

\paragraph{Dialogue Generation.}
\cref{tab:large-exp2_full} provides full experimental results for the dialogue generation task. The results for \gls{ours} showcase the highest scores for all metrics in both cases of rank 64 and rank 4. Despite the conflicting correlations between BLEU and METEOR metrics~\citep{kilickaya-etal-2017-evaluating,agarwal-lavie-2008-meteor}, \gls{ours} achieves comprehensive improvements across all metrics, distinguishing itself from other baselines that fail to achieve such comprehensive enhancements. This proves the efficacy of our multi-objective method in effectively considering multiple metrics with conflicting correlations.

\begin{table*}[t]
\centering
\caption{\textbf{Full Results on Question-Answering Task with \gls{t5}-base.} Results of \gls{ours} and baseline methods with SQuAD2.0 dataset. F1 and EM denote the F1 score and Exact Match, respectively.}
\label{tab:mid_exp2_full}
\begin{small}
\begin{sc}
\begin{tabular}{lcccc}
\toprule
Method & F1 & EM & Avg. \\
\midrule
HPBO (Full) & 78.28 & 73.29 & 75.79\\ 
SWA & 80.31 & 74.85 & 77.58\\
Greedy SWA & 80.63 & \UL{75.44} & \UL{78.03}\\
Learned SWA & \UL{80.65} & 74.23 & 77.44 \\
TWA  & 80.29 & 74.79 & 77.54\\
BOMF (ours) & \BL{80.82} & \BL{75.79} & \BL{78.31}\\
\midrule
HPBO (Freeze)& 78.19 & 73.43 & 75.81 \\
SWA & 81.21 & 75.63 & 78.42\\
Greedy SWA & \UL{81.73} & \UL{76.20} & \UL{78.97}\\
Learned SWA & 81.24 & 75.65 & 78.45\\
TWA  & 81.21 & 75.61 & 78.41\\
BOMF (ours)& \BL{81.82} & \BL{76.21} & \BL{79.01}\\
\bottomrule
\end{tabular}
\end{sc}
\end{small}
\end{table*}

\begin{table*}[t]
\centering
\caption{\textbf{Full Results on Summarization Task with \gls{llama27b}.} Results of \gls{ours} and baseline methods with SAMSum dataset. R1, R2, and RL denote Rouge-1, Rouge-2, and Rouge-L, respectively.}
\label{tab:large_exp1_full}
\begin{small}
\begin{sc}
\begin{tabular}{lcccc}
\toprule
Method & R1 & R2 & RL & Avg. \\ 
\midrule
HPBO (Rank 64) & 52.66 & 28.22 & \UL{44.33} & 41.73\\
SWA & 51.81 & 27.61 & 43.55 & 40.99 \\
Greedy SWA & \BL{53.40} & 28.06 & 43.31 & 41.49 \\
Learned SWA & 52.93 & \BL{28.97} & 44.04 & \UL{41.98}\\
BOMF (ours)& \BL{53.40} & \UL{28.78} & \BL{44.38} & \BL{42.19} \\ 
\midrule
HPBO (Rank 4) & 52.25 & 27.52 & 44.03  & 41.27  \\
SWA & 52.21 & 27.58 & 44.04 & \UL{41.28} \\
Greedy SWA & \UL{53.06} & \BL{28.99} & 44.03 & \BL{42.03} \\
Learned SWA & 52.10 & 26.76 & \UL{44.13} & 41.00  \\
BOMF (ours)& \BL{53.07} & \UL{28.61} & \BL{44.40} & \BL{42.03}  \\

\bottomrule
\end{tabular}
\end{sc}
\end{small}
\end{table*}

\begin{table*}[t!]
\centering
\caption{\textbf{Full Results on Dialogue Generation Task with \gls{llama27b}.} Results of \gls{ours} and baseline methods with E2E dataset. B, M, and RL denote BLEU, METEOR, and Rouge-L, respectively.}
\label{tab:large-exp2_full}
\begin{small}
\begin{sc}
\begin{tabular}{lccccc}
\toprule
Metric & B & M & RL & Avg. \\
\midrule
HPBO (Rank 64) & 63.09 & \UL{80.17} & 65.71 & 69.66  \\
SWA & 62.95 & 80.07 & 65.53 & 69.52\\
Greedy SWA & \UL{63.86} & 80.00 & \UL{66.63} & \UL{70.16}\\
Learned SWA & 63.65 & 79.76 & 66.34 & 69.92 \\
BOMF (ours) & \BL{64.70} & \BL{80.91} & \BL{67.53} & \BL{71.05} \\
\midrule
HPBO (Rank 4) & 63.46 & \UL{81.26} & 67.09 & 70.60 \\
SWA & 63.50 & 81.02 & 67.08 &70.53\\
Greedy SWA & 63.27 & 81.17 & \UL{67.68} & \UL{70.71}\\
Learned SWA & \UL{64.42} & 79.04 & 66.09 & 69.85 \\
BOMF (ours) & \BL{64.81} & \BL{81.28} & \BL{67.70} & \BL{71.26} \\
\bottomrule
\end{tabular}
\end{sc}
\end{small}
\end{table*}
\clearpage
\section{Societal impact}
\label{app:sec:societalimpact}

\gls{ours} does not directly have any positive or negative societal impacts since our algorithm is for fine-tuning and model fusion. However, in the sense of developing \glspl{llm},
we can argue the societal impacts of our work.
On the positive side,
our work can improve the productivity of human beings,
e.g., reduction of repeating tasks,
and discover new scientific knowledge, e.g., artificial intelligence for scientific discovery.
On the other hand,
as negative societal impacts,
fine-tuning \glspl{llm} on downstream tasks can still consume significant compute resources,
which leads to climate change.
Moreover, since our fine-tuning process aims to optimize specific metrics, there can be a potential risk of optimizing towards malicious metrics such as aggressiveness and violence.
Therefore, we should be aware of potential unethical outcomes and consider responsibility in selecting and optimizing these metrics.


\section{Safeguards}
\label{app:sec:safeguard}

We use publicly available benchmarks and open-source models widely recognized in the \gls{llm} research community.
Additionally, we do not release proprietary or new datasets or models that could cause the risk of misuse.
Although our work potentially has a possibility to undertake inherent misuse that is derived from public benchmarks and open-source models,
we think that our method itself does not pose a high risk of misuse.
\newpage
\clearpage
\newpage
\section*{NeurIPS Paper Checklist}

\begin{enumerate}

\item {\bf Claims}
    \item[] Question: Do the main claims made in the abstract and introduction accurately reflect the paper's contributions and scope?
    \item[] Answer: \answerYes{} 
    \item[] Justification: As outlined in \cref{main:sec:introduction}, our approach proposes a model fusion method that leverages \gls{bo} techniques. The empirical results in \cref{main:sec:experiments} demonstrate that \gls{ours} outperforms other baseline methods.

\item {\bf Limitations}
    \item[] Question: Does the paper discuss the limitations of the work performed by the authors?
    \item[] Answer: \answerYes{} 
    \item[] Justification: We discussed the limitations of \gls{ours} in \cref{main:sec:conclusion} and proposed future research to provide theoretical reasoning for our findings.

\item {\bf Theory Assumptions and Proofs}
    \item[] Question: For each theoretical result, does the paper provide the full set of assumptions and a complete (and correct) proof?
    \item[] Answer: \answerNA{} 
    \item[] Justification: We don't have any theoretical results in our paper.

    \item {\bf Experimental Result Reproducibility}
    \item[] Question: Does the paper fully disclose all the information needed to reproduce the main experimental results of the paper to the extent that it affects the main claims and/or conclusions of the paper (regardless of whether the code and data are provided or not)?
    \item[] Answer: \answerYes{} 
    \item[] Justification: We provided a detailed explanation of the algorithm for \gls{ours} in \cref{main:sec:method} and discussed experimental details in \cref{app:sec:experimental_details}. Also, we provide detailed algorithms in \cref{app:sec:algorithm}.

\item {\bf Open access to data and code}
    \item[] Question: Does the paper provide open access to the data and code, with sufficient instructions to faithfully reproduce the main experimental results, as described in supplemental material?
    \item[] Answer: \answerYes{} 
    \item[] Justification: We offer comprehensive experimental settings, including details about the data, in \cref{app:sec:additional_experiment}, and our code is provided in the supplemental material.

\item {\bf Experimental Setting/Details}
    \item[] Question: Does the paper specify all the training and test details (e.g., data splits, hyperparameters, how they were chosen, type of optimizer, etc.) necessary to understand the results?
    \item[] Answer: \answerYes{} 
    \item[] Justification: We provide detailed experimental settings in 
    \cref{app:sec:experimental_details}.

\item {\bf Experiment Statistical Significance}
    \item[] Question: Does the paper report error bars suitably and correctly defined or other appropriate information about the statistical significance of the experiments?
    \item[] Answer: \answerNo{} 
    \item[] Justification: Due to the need to fine-tune Large Language Models across various tasks, conducting experiments multiple times would be excessively computationally expensive in our work.

\item {\bf Experiments Compute Resources}
    \item[] Question: For each experiment, does the paper provide sufficient information on the computer resources (type of compute workers, memory, time of execution) needed to reproduce the experiments?
    \item[] Answer: \answerYes{} 
    \item[] Justification: We provide our computational resources in \cref{app:sec:experimental_details}.
    
\item {\bf Code Of Ethics}
    \item[] Question: Does the research conducted in the paper conform, in every respect, with the NeurIPS Code of Ethics \url{https://neurips.cc/public/EthicsGuidelines}?
    \item[] Answer: \answerYes{} 
    \item[] Justification: Our paper aligns with the NeurIPS Code of Ethics.

\item {\bf Broader Impacts}
    \item[] Question: Does the paper discuss both potential positive societal impacts and negative societal impacts of the work performed?
    \item[] Answer: \answerYes{} 
    \item[] Justification: We discuss the societal impacts regarding \gls{ours} in \cref{app:sec:societalimpact}.
    
\item {\bf Safeguards}
    \item[] Question: Does the paper describe safeguards that have been put in place for responsible release of data or models that have a high risk for misuse (e.g., pretrained language models, image generators, or scraped datasets)?
    \item[] Answer: \answerYes{} 
    \item[] Justification: We discuss the safeguards regarding \gls{ours} in \cref{app:sec:safeguard}.

\item {\bf Licenses for existing assets}
    \item[] Question: Are the creators or original owners of assets (e.g., code, data, models), used in the paper, properly credited and are the license and terms of use explicitly mentioned and properly respected?
    \item[] Answer: \answerYes{} 
    \item[] Justification: We address the code, data, models, and packages utilized in our experiments in~\cref{app:sec:experimental_details}.

\item {\bf New Assets}
    \item[] Question: Are new assets introduced in the paper well documented and is the documentation provided alongside the assets?
    \item[] Answer: \answerNA{} 
    \item[] Justification: We do not provide any new assets.

\item {\bf Crowdsourcing and Research with Human Subjects}
    \item[] Question: For crowdsourcing experiments and research with human subjects, does the paper include the full text of instructions given to participants and screenshots, if applicable, as well as details about compensation (if any)? 
    \item[] Answer: \answerNA{} 
    \item[] Justification: Our paper does not include crowdsourcing or research involving human subjects.

\item {\bf Institutional Review Board (IRB) Approvals or Equivalent for Research with Human Subjects}
    \item[] Question: Does the paper describe potential risks incurred by study participants, whether such risks were disclosed to the subjects, and whether Institutional Review Board (IRB) approvals (or an equivalent approval/review based on the requirements of your country or institution) were obtained?
    \item[] Answer: \answerNA{} 
    \item[] Justification: Our paper does not include crowdsourcing or research involving human subjects.

\end{enumerate}

\end{document}